\PassOptionsToPackage{numbers}{natbib}
\documentclass[twoside,11pt]{article}

\usepackage{blindtext}

\usepackage{lastpage}
\usepackage{xcolor} \usepackage[normalem]{ulem} \usepackage{changebar}

\usepackage[abbrvbib, preprint]{jmlr2e}

\usepackage[utf8]{inputenc}
\usepackage{amsmath}
\usepackage{amsfonts}
\usepackage{amssymb}
\usepackage{dsfont}
\usepackage{xcolor}
\usepackage{hyperref}
\usepackage{mathtools}
\usepackage{pgf}
\usepackage{chngpage}
\usepackage[capitalise, noabbrev]{cleveref}
\usepackage[ruled]{algorithm2e}
\usepackage{longtable}
\usepackage{makecell}
\usepackage{nicefrac}
\usepackage{placeins}

\usepackage{tikz}
\usetikzlibrary{
  graphs,
  graphs.standard
}
\usepackage{relsize}

\tikzset{fontscale/.style = {font=\relsize{#1}}
    }

\usepackage[textwidth=30mm]{todonotes}

\usepackage{bbding}
\usepackage{pifont}

\usepackage{array}
\newcolumntype{x}[1]{>{\centering\arraybackslash\hspace{0pt}}p{#1}}

\newcommand{\ie}{i.e.,\ }
\newcommand{\eg}{e.g.,\ }

\newcommand{\iid}{i.i.d.\ }

\newcommand{\singleregret}{R}

\newcommand{\munderbar}[1]{\text{\underline{$#1$}}}

\newcommand{\bigO}[1]{\mathcal{O}\left(#1\right)}
\newcommand{\kl}[1]{\mathrm{kl}\left(#1\right)}

\newcommand{\btheta}{{\boldsymbol \theta}}

\newcommand{\debug}[1]{#1}

\newcommand{\one}[1]{\mathds{1} \left( #1 \right)}
\newcommand{\card}[1]{\# \left\{ #1 \right\}}
\newcommand{\smallo}[1]{o\left( #1 \right)}
\renewcommand{\mod}[1]{\ (\mathrm{mod} \ #1)}
\DeclareMathOperator*{\argmax}{\arg\!\max}

\newcommand{\pull}{\debug \pi} 
\newcommand{\rew}{\debug r} 
\newcommand{\coll}{\debug \eta} 
\newcommand{\matchset}{\debug \mathcal{M}}
\newcommand{\reg}{\debug R}

\newcommand{\cucb}[1][]{\textsc{CUCB}#1\ }
\newcommand{\escb}[1][]{\textsc{ESCB}#1\ }
\newcommand{\rand}[1][]{\textsc{Rand Orthogonalisation}#1\ }
\newcommand{\musical}[1][]{\textsc{Musical Chairs}#1\ }
\newcommand{\wangalgo}[1][]{\textsc{DPE1}#1\ }
\newcommand{\sicmmab}[1][]{\textsc{SIC-MMAB}#1\ }
\newcommand{\selfish}[1][]{\textsc{Selfish}#1\ }

\jmlrheading{25}{2024}{1-\pageref{LastPage}}{6/22; Revised
10/23}{1/24}{22-0643}{Boursier Etienne and Perchet Vianney}
\ShortHeadings{A Survey on Multi-player Bandits}{Boursier and Perchet}
\firstpageno{1}

\begin{document}

\title{A Survey on Multi-player Bandits}

\author{\name Etienne Boursier \email etienne.boursier1@gmail.com \\
       \addr INRIA, Université Paris Saclay, LMO, Orsay, France
       \AND
       \name Vianney Perchet \email vianney.perchet@normalesup.org \\
       \addr  CREST, ENSAE Paris, Palaiseau, France \\
       CRITEO AI Lab, Paris, France}

\editor{Alexandre Proutiere}

\maketitle

\begin{abstract}%
Due mostly to its application to cognitive radio networks, multiplayer bandits gained a lot of interest in the last decade. A considerable progress has been made on its theoretical aspect. However, the current algorithms are far from applicable and many obstacles remain between these theoretical results and a possible implementation of multiplayer bandits algorithms in real communication networks. This survey contextualizes and organizes the rich multiplayer bandits literature. In light of the existing works, some clear directions for future research appear. We believe that a further study of these different directions might lead to theoretical algorithms adapted to real-world situations.
\end{abstract}

%
%
%
\section{Introduction}

The Multi-Armed Bandits (MAB) problem \citep{thompson33} has been extensively studied in the last decades, probably because of its applications to online recommendation systems, and is one of the most used models of online learning. In its classical version, a single player sequentially chooses an action among a finite set $[K] \coloneqq \lbrace 1, \ldots, K \rbrace$. After pulling the arm $k \in [K]$ at round $t$, the player receives the reward $X_k(t)$, which is drawn from some unknown distribution, of mean denoted by $\mu_k$ in the stochastic setting. Since only the reward of the pulled arm is observed, the player must balance between \textbf{exploration} (acquiring information on the different arms) and \textbf{exploitation} (pulling the seemingly best arm).
This model has known many extensions such as contextual, combinatorial or Lipschitz bandits for example \citep{woodroofe1979one, cesa2012combinatorial, agrawal1995continuum, perchet2013}.

The multiplayer variant of this problem has also known a recent interest, motivated by cognitive radio networks. It considers multiple decentralized players acting on a single Multi-Armed Bandits instance. If several of them pull the same arm at the same time, a \textit{collision} occurs and causes a loss (of the message in the previous example). This leads to a decrease in the received reward and makes the problem much more intricate.

\medskip

\cref{sec:motivation} first presents in more detail the motivations leading to the design of the multiplayer bandits model. The most classical version of multiplayer bandits is then described in \cref{sec:base}, along with a first base study including the centralized case and a lower bound of the incurred regret.
\cref{sec:reaching} presents the different results known for this model. In particular, collisions can be abusively used to reach regret bounds similar to the centralized case.
\cref{sec:realistic} presents several practical considerations that can be added to the model, which might lead to more natural algorithms. Finally, \cref{sec:alternateproblems} mentions the multi-agent bandits, competing bandits and dynamic queuing systems problems, which all bear similarities with multiplayer bandits, either in the model or in the used algorithms.  \cref{sec:banditsintro} introduces single player stochastic MAB and the most common algorithms. \cref{table1,table2} in \cref{app:table} summarize the main results presented in this survey.

\section[Motivation for Cognitive Radio Networks]{Motivation for cognitive radio networks\,\footnote{We are grateful to Christophe Moy for his precious feedback on the application of multiplayer bandits to cognitive radio networks.}}\label{sec:motivation}

The problem of multiplayer bandits is mostly motivated by its applications to cognitive radio, whose paradigm has been first proposed by \citet{mitola1999} and can be defined as a radio capable of learning its environment and choosing dynamically the best configuration for transmission. The definition of best configuration depends on the precise objective and is often considered as the maximal bandwidth usage rate \citep{haykin2005cognitive}. 
Cognitive radios attract a lot of attention due to their numerous potential applications, such as Internet of Things \citep{shah2013cognitive}, smart vehicles \citep{di2012smart}, public safety communication \citep{ferrus2012public} and civil aviation \citep{zheng2023overview}. 
Yet, cognitive radios are still at a development stage. Many technical issues remain to be solved before a possible use of cognitive radios in real world applications.  
In particular, improving the following features are crucial for a successful implementation of cognitive radios: spectrum awareness and exploitation \citep{sharma2015cognitive} and security \citep{attar2012}. 
We refer to \citep{marinho2012, garhwal2012} for surveys on different research directions on cognitive radios. 

The most challenging approach to cognitive radio is \textbf{Opportunistic Spectrum Access} (OSA). Consider a spectrum partitioned into licensed bands. For each band, some designated Primary User (PUs), who pays a license, is given preferential access. 
In practice, many of these bands remain largely unused by PUs. OSA then aims at maximizing the spectrum usage by giving Secondary Users (SUs) the possibility to access channels  left free by the PUs. Assuming the SUs are equipped with a \textit{spectrum sensing} capacity, they can first sense the presence of a PU on a channel to give priority to PUs. If no PU is using the channel, SUs can transmit on this channel. 
Such devices yet have limited capabilities; especially, they proceed in a decentralized network and cannot sense different channels simultaneously. This last restriction justifies the bandit feedback assumed in formal models.

However, the previous sensing procedure is not perfect in practice: SUs might wrongly detect the absence of a PU on a channel, \eg because of the hidden node problem \citep{sahai2004some}.
Moreover, for other applications such as \textbf{Internet of Things} (IoT) networks, the devices have even lower power capabilities and there is no more licensed band. As a consequence, all devices behave as SUs (no PU) and do not require to, or cannot,  sense the presence of another user before transmitting. These devices can still perform some form of learning: if the base station sends back downlink messages, the devices can determine afterward whether their transmission was successful, \ie whether they received an \textit{acknowledgement}. As a consequence, similar learning models can be used for many different communication networks, and only differ from minor assumptions such as the received feedback (see \cref{sec:model}).

Depending on the application, the user can also choose dynamically other transmission parameters,  to maximise its throughput. For example in the case of Wi-Fi networks, the user can choose the transmission mode and rate \citep{combes2014dynamic}.

\medskip

Using a Multi-Armed bandits model to improve spectrum exploitation for cognitive radios was first suggested by \citet{jouini2009,jouini2010,liu2008}. In these first attempts in formalizing the problem, a single SU (player) repeatedly chooses among a choice of $K$ arms (\ie channels or any tunable transmission parameter) for transmission. The success of transmission is a random variable $X_k(t) \in \{0, 1 \}$, where the sequence $(X_k(t))_{t}$ can be \iid (stochastic model) or a Markov chain for instance. 
In particular the Gilbert-Elliot model, with two states, is often considered as a good approximation of PUs activity \citep{tekin2011online}. 
A successful transmission corresponds then to $X_k = 1$,  against $X_k=0$ if transmission failed, \eg the channel was occupied by a PU. The goal of the SU is then to maximize its number of transmitted messages, or in bandit terms, to minimize its regret. In real networks, devices observe whether their message is successfully transmitted using acknowledgement messages, but this aspect is disregarded here for the sake of simplicity.

Shortly after, \citet{liu2010} extended this model to multiple users, taking into account the interaction between SUs in communication networks. The problem becomes more intricate as SUs interfere when transmitting on the same channel. The event of multiple SUs simultaneously using the same channel is called a \textit{collision} in the literature.

\medskip

Different Proof-of-Concepts, simulating networks in laboratory conditions, later justified the use of Reinforcement Learning, and more particularly Multi-Armed bandits model, for OSA \citep{robert2014,toldov2016thompson,kumar2018b}, Wi-Fi networks\footnote{Instead of the channel, the devices here adaptively choose the transmission rate and mode.} \citep{combes2018optimal} and  IoT networks \citep{moy2019decentralized}. 

While OSA currently remains a futuristic application, a real IoT deployment with bandits algorithms has already proven useful  in a LoRaWAN network \citep{moy2019decentralized}. In this work, \citet{moy2019decentralized} implemented a \textsc{UCB} algorithm on the device side in a real LoRa network, in the city of Rennes, France. The device dynamically chooses, with a duty cycle of $1$\% between three channels in the European ISM band at $868.1$ MHz, $868.3$ MHz and $868.5$ MHz. These channels correspond to the authorized frequencies in France. To enable learning from the device side, the application server sends back an acknowledgement to the device in case of successful transmission. 
In only $129$ transmissions, the learning device already outperforms by a factor $2$ (in transmission rate) devices selecting their channel uniformly at random. 

All these advanced experimentations demonstrate the potential utility of bandit policies for enhanced cognitive radio networks. We refer to \citep{jouini2012, besson2019b} for more details on the link between OSA, IoT and Multi-Armed bandits.

\medskip

Besides the application to cognitive radios, which is at the origin of the multiplayer bandits problem, the algorithmic techniques and proof ideas developed for multiplayer bandits can also be adapted to other multi-agent sequential learning problems. Such other problems are mentioned in \cref{sec:alternateproblems} and include applications to matching markets and packet routing in servers.

\section{Baseline Results} \label{sec:base}

This section describes the classical model of multiplayer bandits, with several variations of observation and arm means setting, and gives first results (derived from the centralized case), as well as notations used along this survey. Harder, more realistic variations are discussed in \cref{sec:realistic}. 

For completeness, \cref{sec:banditsintro} provides a short introduction to the classical (single player) stochastic MAB problem as well as the common algorithms and results known for this problem. Algorithms described in the remaining of the survey are based on these common algorithms and include \textsc{UCB} (Upper Confidence Bound), $\varepsilon$-Greedy, \textsc{ETC} (Explore-Then-Commit).
%
%

In this survey, we write $f(T)=\bigO{g(T)}$ if there exists a universal constant $c\in \mathbb{R}$ such that for any $T\geq 1$, $f(T)\leq c g(T)$. The constant $c$ must not depend on any problem parameters.
Conversely, we note $f(T)=\Omega(g(T))$ if $g(T)=\bigO{f(T)}$ and, furthermore, $f(T)=\smallo{g(T)}$ if $\lim_{T\to\infty} \frac{f(T)}{g(T)} = 0$.

\subsection{Model} \label{sec:model}

Consider a bandit problem with $M$ players and $K$ arms, where $M\leq K$. To each arm-player pair is associated an \iid sequence of rewards $(X_k^m(t))_{t \in [T]}$, where $X_k^m$ follows a distribution in $[0, 1]$ of mean $\mu_k^m$.
At each round $t \in [T] \coloneqq \{1, \ldots, T\}$, all players pull simultaneously an arm in $[K]$. 
We denote by $\pull^m(t)$ the arm pulled by player $m$ at time $t$, who receives the individual reward
\begin{equation*}
\rew^m(t) \coloneqq X_{\pull^m(t)}^m(t) \cdot (1-\coll_{\pull^m(t)}(t)),\end{equation*}
where $\coll_{k}(t) = \one{\card{m \in [M] \mid \pull^m(t) = k} > 1}$ is the collision indicator, and $\# A$ denotes the cardinality of a set $A$.
The players are assumed to know the horizon $T$, even though this is not crucial \citep{degenne2016anytime},   and use a common numbering of the arms.

\medskip

A \textit{matching} $\pi \in \matchset$ is an assignment of players to arms, \ie mathematically, is a one to one function $\pi : [M] \rightarrow [K]$. The (expected) utility of a matching is then defined as
\begin{equation*}
U(\pi) \coloneqq \sum_{m=1}^M \mu_{\pi(m)}^m.
\end{equation*}
The performance of an algorithm is measured in terms of (pseudo-)regret, which is the difference between the maximal expected reward and the algorithm cumulative reward:
\begin{equation*}
\reg(T) \coloneqq T U^* - \sum_{t=1}^T \sum_{m=1}^M \mu_{\pull^m(t)}^m \cdot (1-\coll_{\pull^m(t)}(t)),
\end{equation*}
where $U^* = \max_{\pi \in \matchset}U(\pi)$ is the maximal realizable utility. The problem difficulty is related to the suboptimality gap $\Delta$ where
\begin{equation*}\Delta \coloneqq U^* - \max\left\{U(\pi) \mid \pi\in\matchset,\ U(\pi) < U^* \right\}.\end{equation*}
%
In contrast to the classical bandits problem where only the received reward can be observed at each time step, multiplayer settings might differ in the received feedback at each step, which leads to at least four different settings\footnote{\citet{bubeck2020b} also consider a fifth setting where only $X^m_{\pull^m(t)}(t)$ is observed in order to completely ignore collision information.}, described in \cref{table:obs_settings} below.

\begin{table}[h]
\begin{adjustwidth}{-50pt}{-50pt}\centering
\setlength{\extrarowheight}{4pt}
\setlength\tabcolsep{3pt}
 \begin{tabular}{|x{80pt} || x{70pt}|x{70pt}|x{70pt}|x{60pt}|} 
 \hline
 \textbf{Setting} & Full sensing &  Statistic sensing & Collision sensing  & No-sensing \\[4pt]
 \hline
 \textbf{Feedback} &  $\coll_{\pull^m(t)}(t)$ and $X^m_{\pull^m(t)}(t)$ & $X^m_{\pull^m(t)}(t)$ and $\rew^m(t)$ & $\coll_{\pull^m(t)}(t)$ and $\rew^m(t)$ & $\rew^m(t)$ \\[8pt] \hline
\end{tabular}
\end{adjustwidth}
\caption{\label{table:obs_settings}Different observation settings considered. \textit{Feedback} represents the observation of player $m$ after round $t$.}
\end{table}

The different settings can be motivated by different applications, or purely for theoretical purposes. For example, statistic sensing models the OSA problem, where SUs first sense the presence of a PU before transmitting on the channel, assuming there is no sensing failure (\eg because of hidden nodes). On the other hand, no-sensing can model IoT networks as explained in \cref{sec:motivation}. 
The no-sensing setting is obliviously the hardest one, since a $0$ reward can either correspond to a low channel quality or a collision with another player.

\medskip

The above description corresponds to the \textit{heterogeneous} setting, where the arm means differ among the players. In practice, the quality of a channel might vary among players, notably because of propagation problems such as frequency selective fading and path loss. In the following, the easier \textit{homogeneous} setting is also considered, where the arm means are common to all players: $\mu_k^m = \mu_k$ for all $(m,k) \in [M]\times [K]$.  If $\mu_{(k)}$ denotes the $k$-th largest mean, \ie $\mu_{(1)} \geq \mu_{(2)} \ldots \geq \mu_{(K)}$, the maximal expected reward is given by
$$
\max_{\pi \in \matchset} U(\pi) = \sum_{k=1}^M \mu_{(k)},
$$
which largely facilitates the learning problem.

The statistics $(X_k^m(t))$ can be either common or different between players in the literature. In the following, we consider by default common statistics between players and precise when otherwise. Note that this has no influence in both collision and no-sensing settings.

\subsection{Centralized Case}\label{sec:centralized}

To set baseline results,  we consider first, in this section, the easier centralized model, where all players in the game described in \cref{sec:model} are controlled by a common central agent. It becomes trivial for this central agent to avoid collisions between players as she unilaterally decides the arms they pull. The difficulty thus only is determining the optimal matching~$\pi$ in this simplified setting.

\medskip

\textbf{Bandits with multiple plays.} In the homogeneous setting where the arm means do not vary across players, the centralized case reduces to bandits with multiple plays, where a single player has to pull $M$ arms among a set of $K$ arms at each round. \citet{anantharam1987a} introduced this problem long before multiplayer bandits and provided an asymptotic lower bound for this problem, given by \cref{thm:lowerboundhomo} in \cref{sec:lowerbound}. 
They also provided an optimal algorithm, asymptotically reaching this exact regret bound.

\medskip

\textbf{Combinatorial bandits.} More generally, multiple plays bandits as well as the heterogeneous centralized setting are particular instances of combinatorial bandits \citep{gai2012}, where the central agent plays an action (representing several arms) $a \in \mathcal{A}$ and receives $r(\pmb{\mu}, a)$ for reward. We here consider the simple case of linear reward $r(\pmb{\mu}, a) = \sum_{k \in a}\mu_k$.

In the homogeneous case, $\mathcal{A}$ is all the subsets of $[K]$ of size $M$. 
In the heterogeneous case, however, $MK$ arms are considered instead of $K$ (one arm per pair $(m,k)$) and $\mathcal{A}$ is the set of matchings between players and arms.

\citet{chen2013} proposed the \cucb algorithm, which yields a $\bigO{\frac{M^2 K}{\Delta}\log(T)}$ regret in the heterogeneous setting \citep{kveton2015}. While \cucb performs well for any correlation between the arms, \citet{combes2015} leverage the independence of arms with \escb to reach a $\bigO{\frac{\log^2(M) M K}{\Delta}\log(T)}$ regret in this specific setting. \escb however suffers from computational inefficiencies in general, as it requires computing upper confidence bounds for any (meta-)action. The number of such actions indeed scales combinatorially with the number of arms and players. Thompson Sampling strategies remedy this problem, while still having $\bigO{\frac{\log^2(M) MK}{\Delta}\log(T)}$ regret for independent arms \citep{wang2018}. \citet{cuvelier2021asymptotically} proposed a computationally efficient, asymptotically optimal algorithm for many combinatorial bandits problems with independent arms, including the centralized heterogeneous one. The optimal regret bound is yet not explicit, but characterized as the minimum of some optimization problem. 
\citet{degenne2016b} and \citet{perrault2020} respectively extended \escb and combinatorial TS for the intermediate case of any fixed correlation between the arms.


\subsection{Lower Bound}\label{sec:lowerbound}

This section describes the different lower bounds known in multiplayer bandits, which are derived from the centralized case.

As mentioned in \cref{sec:centralized}, \citet{anantharam1987a} provided a lower bound for the centralized homogeneous setting. This setting is obviously easier than the decentralized homogeneous multiplayer problem so this bound also holds for the latter.

\begin{definition}
We call an algorithm uniformly good if for every parameter configuration $\pmb{\mu}, K, M$, it yields for every $\alpha>0$ that $\reg(T) = \smallo{T^\alpha}$.
\end{definition}
\begin{theorem}[\citealt{anantharam1987a}]\label{thm:lowerboundhomo}
For any uniformly good algorithm and all instances of homogeneous multiplayer bandits with $\mu_{(1)} > \ldots > \mu_{(K)}$,
\begin{equation*}
\liminf_{T \to \infty} \frac{\reg(T)}{\log(T)} \geq \sum_{k>M} \frac{\mu_{(M)}-\mu_{(k)}}{\kl{\mu_{(M)}, \mu_{(k)}}},
\end{equation*}
where $\kl{p,q} = p\log\left(\frac{p}{q}\right) + (1-p)\log\left( \frac{1-p}{1-q}\right)$.
\end{theorem}
\citet{combes2015} proved a lower bound for general combinatorial bandits, 
determined as the solution of an optimization problem. Luckily for the specific case of matchings, its value is simplified. Especially, for some problem instances of the heterogeneous setting, any uniformly good algorithm regret is $\Omega\left(\frac{KM}{\Delta}\log(T)\right)$. 
%
In the centralized case, studying the heterogeneous setting is already more intricate than the homogeneous one. This gap also holds when considering decentralized algorithms as shown in the following sections.

It was first thought that the decentralized problem was harder than the centralized one and that an additional $M$ factor, the number of players, would appear in the regret bounds of any decentralized algorithm \citep{liu2010, besson2018}. This actually only holds if the players do not use any information from the collisions incurred with other players \citep{besson2019a}; but as soon as the players use this information, only the centralized bound holds.

\section{Reaching Centralized Optimal Regret}\label{sec:reaching}

We shall consider from now on the decentralized multiplayer bandits problem. This section shows how the collision information has been used in the literature, from a coordination to a communication tool between players, until reaching a near centralized performance in theory. In the following, all algorithms are written from the point of view of a single player to highlight their decentralized aspect.

\subsection{Coordination Routines}\label{sec:limited}

The main challenge of multiplayer bandits comes from additional loss due to collisions between players. The players cannot try solely to minimize their individual regret without considering the multiplayer environment, as they would encounter a large number of collisions.
In this direction, \citet{besson2018} studied the behavior of the \selfish algorithm, where players individually follow a UCB algorithm. Although it yields good empirical results on average, players appear to incur a linear regret in some runs. \citet{boursier2018} later proved that \selfish yields a regret scaling linearly with $T$ for machines with infinite precision (\ie that can distinguish an irrational number from an arbitrarily close rational number). It yet remains to be proved for machines with finite bit representation of real numbers.

\medskip 

The first attempts at proposing algorithms for multiplayer bandits considered the homogeneous setting, as well as the existence of a pre-agreement between players \citep{anandkumar2010}. If players are assumed to have distinct ranks $j \in [M]$ beforehand (\ie unique IDs), player $j$ then just focuses on pulling the $j$-th best arm. \citet{anandkumar2010} proposed the first algorithm in this line, using an $\varepsilon$-greedy strategy. Instead of targeting the $j$-th best arm, players can instead rotate in a delayed fashion on the $M$-best arms. For example, when player $1$ targets the $k$-th best arm, player $j$ targets the $k_j$-th best arm where $k_j = k + j -1 \mod{M}$. \citet{liu2010} used a similar UCB-strategy with rotation among players.

\medskip

This kind of pre-agreement among players is however undesirable, and many works instead suggested that the players use collision information for coordination in the absence of pre-agreement. A significant objective of multiplayer bandits is then to \textit{orthogonalize} players, \ie reach a state where all players pull different arms and no collision happens. 

A first routine for orthogonalization, called \rand is given by Algorithm~\ref{alg:randortho} below. Each player pulls an arm uniformly at random among some set (the $M$-best arms or all arms for instance). If she encounters no collision, she continues pulling this arm until receiving a collision. As soon as she encounters a collision, she then restarts sampling uniformly at random. After some time, all players end up pulling different arms with high probability. \citet{anandkumar2011} and \citet{liu2010} used this routine when selecting an arm among the set of the $M$ largest UCB indexes to limit the number of collisions between players.

\citet{avner2014} used a related procedure with an $\varepsilon$-greedy algorithm, but instead of systematically resampling after a collision, players resample only with a small probability $p$. When a player gives up an arm by resampling after colliding on it, she marks it as occupied and stops pulling it for a long time.

\begin{figure*}[htp]\begin{adjustwidth}{-0.2cm}{-0.2cm}
\centering
\begin{minipage}{0.47\textwidth}

\begin{algorithm}[H]\label{alg:randortho}
\DontPrintSemicolon
\SetKwInOut{Input}{input}
\Input{time $T_0$, set $\mathcal{S}$}
\caption{\rand}
$\eta_k(0) \gets 1$ \;
\For{$t \in [T_0]$}{
	\uIf{$\eta_k(t-1)=1$}{Sample $k$ uniformly in $\mathcal{S}$}
	Pull arm $k$ \;
}

\end{algorithm}
   \end{minipage} \hfill   
   \begin{minipage}{0.47\textwidth}
\begin{algorithm}[H]\label{alg:musicalchair}
\DontPrintSemicolon
\SetKwInOut{Input}{input}
\Input{time $T_0$, set $\mathcal{S}$}
$\mathrm{stay} \gets \mathrm{False}$ \;
\For{$t \in [T_0]$}{
	\uIf{$\mathrm{not(stay)}$}{Sample $k$ uniformly in $\mathcal{S}$}
	Pull arm $k$ \;
	\uIf{$\eta_k(t)=0$}{$\mathrm{stay} \gets \mathrm{True}$} 
}
\caption{\musical}
		\end{algorithm}
   \end{minipage}
   \end{adjustwidth}
   \end{figure*}

\citet{rosenski2016} later introduced a faster routine for orthogonalization, \musical described in Algorithm~\ref{alg:musicalchair}. Players sample at random as \rand[,] but as soon as a player encounters no collision, she remains idle on this arm until the end of the procedure, even if she encounters new collisions afterward. This routine is faster since players do not restart each time they encounter a new collision. 

\citet{rosenski2016} used this routine with a simple Explore-then-Commit (ETC) algorithm. Players first pull all arms $\log(T)/\Delta^2$ times so that they know the $M$ best arms afterward while sampling uniformly at random. Players then play musical chairs on the set of $M$ best arms and remain idle on their attributed arm until the end. \citet{joshi2018} proposed a similar strategy but used \musical directly at the beginning of the algorithm so that players rotate over the arms even during the exploration, avoiding additional collisions.

\citet{besson2018} adapted both orthogonalization routines with a UCB strategy. They show that even in the statistic sensing setting where collisions are not directly observed, these routines can be used for orthogonalization. 
\citet{lugosi2018} even used \musical with no-sensing, but require the knowledge of a lower bound of $\mu_{(M)}$. Indeed, for arbitrarily small means, observing only zeros on an arm might not be due to collisions. 
While the ETC algorithm proposed by \citet{rosenski2016} assumes the knowledge of $\Delta$, \citet{lugosi2018} remove this assumption by instead using a Successive Accept and Reject (SAR\footnote{SAR is an extension of the ETC algorithm to the multiple plays bandits problem.}) algorithm \citep{bubeck2013} with epochs of increasing sizes. At the end of each epoch, players eliminate the arms appearing suboptimal and accept arms appearing optimal. The remaining arms still have to be explored in the next phases. To avoid collisions on the remaining arms, players proceed to \musical at the beginning of each new epoch.

\citet{kumar2018} proposed an ETC strategy based on \musical[.] However, they do not require the knowledge of $M$ when assigning the $M$ best arms to players, but instead, use a scheme where players improve their currently assigned arm when possible.

\medskip

With a few exceptions \citep{avner2014, kumar2018}, the presented algorithms require the knowledge of the number of players $M$ at some point, as the players must exactly target the $M$ best arms. While some of them assume $M$ to be a priori known, others estimate it. Especially, uniform sampling rules are useful here, since the number of players can be deduced from the collision probability \citep{anandkumar2011, rosenski2016, lugosi2018}. Indeed, assume all players are sampling uniformly at random among all arms. The probability to collide for a player at each round is exactly $1-(1-1/K)^{M-1}$. If this probability is estimated tightly enough, the number of players is then exactly estimated.

\citet{joshi2018} proposed another routine to estimate $M$. If all players except one are orthogonalized and rotate over the $K$ arms while the remaining one stays idle on a single arm, the number of collisions observed by this player during a window of $K$ rounds is then $M-1$. \citet{joshi2018} also proposed this routine with no-sensing, in which case some lower bound on $\mu$ has to be known similarly to \citep{lugosi2018}.

\bigskip

\textbf{Heterogeneous setting.} All the previous algorithms reach a sublinear regret in the homogeneous setting. Reaching the optimal matching in the heterogeneous setting is yet much harder with decentralized algorithms and the first works on this topic only proposed solutions reaching Pareto optimal matchings. A matching is Pareto optimal if no two players can exchange their assigned arms (or choose a free arm) while both receiving the same or a larger reward. 

\citet{avner2018} and \citet{darak2019} both proposed algorithms with similar ideas to reach a Pareto optimal matching. First, the players are orthogonalized. The time is then divided in several windows. In each window, with a small probability $p$, a player becomes a leader. The leader then suggests switching with the player pulling her currently preferred arm (in UCB index). If this player refuses, the leader then tries to switch for her second preferred arm, and so on. This algorithm thus finally reaches a Pareto optimal matching when all arms are well estimated.

Pareto optimal matchings yet do not guarantee a large social welfare. \cref{fig:PoAmatching} below gives an example of Pareto optimal matching (in green) which yields an arbitrarily small utility when $\varepsilon\to 0$, while the welfare maximising matching (in grey, dashed) yields a utility $1$. In Game Theory lingo, this corresponds to a game with a \textit{Price of Anarchy} $\frac{1}{2\varepsilon}$ \citep{koutsoupias1999worst}.

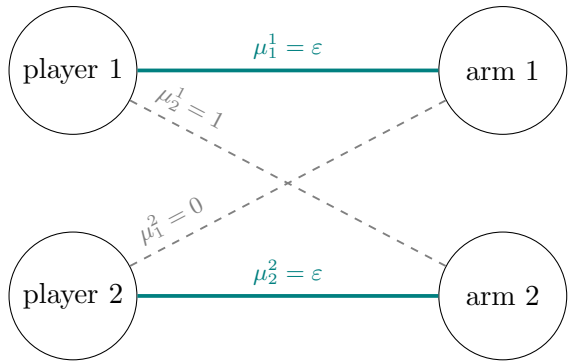
\begin{figure}[ht]\centering
\begin{tikzpicture}
\centering
   \graph[nodes={draw, circle, inner sep=0pt, minimum size=1.7cm},
           branch down=3cm,
           grow left sep=4cm] 
           {subgraph I_nm [W={player 1, player 2},V={arm 1, arm 2}];
};
\draw[dashed, line width=0.25mm,color=gray,fontscale=-1, sloped] (player 1) -- node[above left=0cm and 1cm] {$\mu_2^1=1$} (arm 2) ;
\draw[dashed, line width=0.25mm,color=gray,fontscale=-1, sloped] (player 2) -- node[above left=0cm and 1cm] {$\mu_1^2=0$} (arm 1) ;
\draw[line width=0.5mm,color=green!50!blue,fontscale=-1] (player 1) -- node[above] {$\mu_1^1=\varepsilon$} (arm 1) ;
\draw[line width=0.5mm,color=green!50!blue,fontscale=-1] (player 2) -- node[above] {$\mu_2^2=\varepsilon$} (arm 2) ;
\end{tikzpicture}
\caption{\label{fig:PoAmatching}Example of \textit{bad} Pareto optimal matching.}
\end{figure}

\subsection{Enhancing Communication}\label{sec:fullcomm}

Most of the literature described in \cref{sec:limited} used collision information as a tool for coordination, \ie to avoid collisions between players. Yet,	a richer level of information seems required to reach the optimal allocation in the heterogeneous case. Indeed, the sole knowledge of other players' preferences order is not sufficient to compute the best matching between players and arms. Instead, players need to be able to exchange information about their arms means.

For this purpose, \citet{kalathil2014} assumed that players were able to send real numbers to each other at some rounds. The players can then proceed to a Bertsekas Auction algorithm \citep{bertsekas1992} by bidding on arms to end up with the optimal matching. The algorithm proceeds in epochs of doubling size. Each epoch starts with a decision phase, where players bid according to UCB indexes of their arms. After this phase, players are attributed via $\varepsilon$-optimal matching for these indexes and pull this matching for the whole exploitation phase. This algorithm was later improved and adapted to ETC and TS strategies \citep{nayyar2016}.

Although these works provide the first algorithms with a sublinear regret in the heterogeneous setting, they assume undesirable communication possibilities between players. Actually, this kind of communication is possible through collision observations, when smartly used. In the following of this section, we consider the collision sensing setting if not specified, so that a collision is systematically detected.

\subsubsection{Communication via Markov Chains.}

\citet{bistritz2020b} adapted a Markov chain dynamic \citep{marden2014} for multiplayer bandits to attribute the best matching to players.
Here as well, the algorithm proceeds in epochs of increasing sizes. Each epoch is divided in an exploration phase where players estimate the arm means; a Game of Thrones (GoT) phase, which follows a Markov chain dynamic to attribute the best-estimated matching to players; and an exploitation phase where players pull the matching attributed by the GoT phase. This algorithm reaches a regret scaling with the horizon as $\log^{1+\delta}(T)$ for any choice of parameter $\delta>0$ in the heterogeneous setting (the regret obviously also depends in the parameters $\delta,\mu,K$ and $M$). 

\medskip

The main interest of the algorithm comes from the GoT phase, described in Algorithm~\ref{alg:GoT}, which allows the players to determine the best matching using only collision information. In this phase, players follow a decentralized game, where they tend to explore more when discontent (state $D$) and still explore with a small probability when content (state $C$). When the routine parameters $\varepsilon$ and $c$ are well-chosen, players visit more often the best matching according to the estimated means $\hat{\mu}_k^j$ so far. Each player, while content, pulls her assigned arm in the optimal matching most often. This phase thus allows estimating the optimal matching between arms and players as proved by \citet{bistritz2020b}. 

\begin{algorithm}[h]
\DontPrintSemicolon
\SetKwInOut{Input}{input}
\Input{time $T_0$, starting arm $\bar{a}_t$, player $j$, parameters $\varepsilon$ and $c$}
\caption{\label{alg:GoT}Game of Thrones subroutine}
$S_t \gets C$; $u_{\max} \gets \max_{k \in [K]} \hat{\mu}_k^j$\;
\For{$t = 1, \ldots, T_0$}{
\lIf{$S_t = C$}{pull $k$ with probability $\begin{cases}1-\varepsilon^c \text{ if } k=\bar{a}_t \\ \varepsilon^c/(K-1) \text{ otherwise}\end{cases}$}
\lElse{pull $k$ with probability $1/K$}
\uIf{$k\neq\bar{a}_t$ or $\eta_k(t) =0$ or $S_t=D$}{
$\bar{a}_t, S_t \gets \begin{cases} k, C \text{ with probability } \frac{\hat{\mu}_k^j \, \eta_k(t)}{u_{ \max}}\varepsilon^{u_{\max}-\hat{\mu}_k^j \eta_k(t)} \\ k, D \text{ otherwise} \end{cases}$}
}
\end{algorithm}
		
\citet{youssef2020} extended this algorithm to the multiple plays setting, where each player can pull several arms at each round.

This routine elegantly assigns the optimal matching to players. However, it suffers from a large dependency in other problem parameters than $T$, as the GoT phase requires the Markov chain to reach its stationary distribution. Also, the algorithm requires a good tuning of the GoT parameters $\varepsilon$ and $c$, which depend on the suboptimality gap $\Delta$. 

\subsubsection{Collision Information as Bits.} In a different work, \citet{boursier2018} suggested with \sicmmab algorithm that the collision information $\eta_k(t)$ can be interpreted as a bit sent from a player $i$ to a player $j$, if they previously agreed that at this time, player $i$ was sending a message to player $j$. For example, a collision represents a $1$ bit, while no collision a $0$ bit.

Such an agreement is possible if the algorithm is well designed and different ranks in $[M]$ are assigned to the players. These ranks are here assigned using an initialization procedure based on Musical chairs \citep{boursier2018}, which also quickly estimates the number of players $M$. This initialization procedure, later improved by \citet{wang2020}, allows to reach the pre-agreement\footnote{In the more difficult settings considered in \cref{sec:realistic}, whether this pre-agreement can be reached at a negligible cost remains an open problem.} mentioned in \cref{sec:limited} after a time of order $\bigO{K^2M}$.

\medskip

\textbf{Homogeneous setting.}
After this initialization, the SAR-based algorithm of \citet{boursier2018} runs epochs of doubling size. Each epoch is divided in an exploration phase, where players pull all accepted arms and arms to explore. In the communication phase, players then send to each other their empirical means (truncated up to a small error) in binary, using collision information as bits. From then, players have shared all their statistics and can accept/eliminate in common the optimal/suboptimal arms. These epochs go on until $M$ arms have been accepted. The players then pull these arms until $T$, without colliding.

Note that the communication regret of \sicmmab can directly be improved using a leader who gathers all the information and communicates the arms to pull to other players \citep{boursier2019}.

\medskip

As the players share their statistics altogether, \citet{boursier2018} showed that the centralized lower bound was achievable with decentralization up to constant factors, contradicting first intuitions. Besides the initialization procedure of \sicmmab[,] \citet{wang2020} also improved the regret due to exploration. In particular, they proposed to elect a leader, who is the only one to explore, while other players greedily exploit the best empirical arms. When the set of best empirical arms changes (which only happens $\bigO{\frac{K^{\nicefrac{3}{2}}M^2}{\Delta^2}}$ times), the leader communicates to the other players the new set of best empirical arms. 
Their algorithm exactly matches the theoretical lower bound for the homogeneous setting.
\begin{theorem}[\citealt{wang2020}]\label{thm:upperboundhomo}
\wangalgo algorithm, in the homogeneous with collision sensing setting such that $\mu_{(M)}>\mu_{(M+1)}$, has an asymptotic regret bounded as
\begin{equation*}
\limsup_{T \to \infty} \frac{\reg(T)}{\log(T)} \leq \sum_{k>M} \frac{\mu_{(M)}-\mu_{(k)}}{\kl{\mu_{(M)}, \mu_{(k)}}}.
\end{equation*}
\end{theorem}
%
A similar leader-followers idea was also proposed by  \citet{verma2019}.
\citet{shi2020a} extended the \sicmmab algorithm to the no-sensing case using \textit{Z-channel coding}. It yet requires the knowledge of a lower bound of the arm means $\mu_{\min}$. Indeed, while a collision is detected in a single round with collision sensing, it can be detected with high probability in $\frac{\log(T)}{\mu_{\min}}$ rounds without sensing. The suboptimality gap $\Delta$ is also assumed to be known here, to fix the number of sent bits at each epoch (while $p$ bits are sent after the epoch $p$ in \sicmmab[).]

\medskip

\citet{huang2021} overcome this issue by proposing a no-sensing algorithm without additional knowledge of problem parameters. In particular, it neither requires prior knowledge of $\mu_{\min}$ nor has a regret scaling with $\frac{1}{\mu_{\min}}$. Such a result is made possible by electing a good arm before the initialization. 
The players indeed start the algorithm with a procedure, such that afterward, with high probability, they have elected an arm $\bar{k}$, which is the same for all players. Moreover, players have a common lower bound of $\mu_{\bar{k}}$, which is of the same order as $\mu_{(1)}$. Thanks to this, the players can then send information on this arm in $\frac{\log(T)}{\mu_{(1)}}$ rounds. This then makes the communication regret independent from the means $\mu_k$, since the regret generated by a collision is at most $\mu_{(1)}$. After electing this good arm, players then follow an algorithm similar to \citet{shi2020a}, with a few modifications to ensure that players only communicate on the good arm~$\bar{k}$.
Yet the communication cost remains large, \ie of order $KM^2\log(T)\log\left(\frac{1}{\Delta}\right)^2$, as sending a bit requires a time of order $\log(T)$ here. Although this term is often smaller than the exploration (centralized) regret, it can be much larger for some problem parameters. Reducing this communication cost thus remains left for future work. 

\citet{pacchiano2021instance} also proposed a no-sensing algorithm without prior knowledge of $\mu$. Although it proceeds to much fewer communication rounds than \citet{huang2021}, it leads to a larger regret bound and requires a pre-agreement on the players' ranks.

\medskip

\textbf{Heterogeneous setting.} The idea of considering collision information as bits sent between players can also be used in the heterogeneous setting. Indeed, this allows the players to share their estimated arm means, and then compute the optimal matching. 
If the suboptimality gap $\Delta$ is known, then a natural algorithm \citep{magesh2019a} estimates all the arms with a precision $\Delta/(2M)$. All players then communicate their estimations, compute the optimal matching and stick to it until $T$. 

When $\Delta$ is unknown, \citet{tibrewal2019} proposed an ETC algorithm, with epochs of increasing sizes. Each epoch consists in an exploration phase where players pull all arms; a communication phase where players communicate their estimated means; and an exploitation phase where players pull the best-estimated matching.

\medskip

\citet{boursier2019} extended \sicmmab to the heterogeneous setting, besides improving its communication protocol with the leader/follower scheme mentioned above. The main difficulty is that players have to explore matchings here. However, exploring all matchings leads to a combinatorial regret and computational complexity of the algorithm. Players instead explore arm-player pairs and the SAR procedure thus accept/reject pairs that are sure to be involved/missing in the optimal matching. 

With a unique optimal matching, similarly to \sicmmab[,] exploration ends at some point and players start exploiting the optimal matching. Besides holding only with a unique optimal matching, this bound incurs an additional $M$ factor with respect to centralized algorithms such as \cucb[.]
In the case of several optimal matchings (in opposition to works mentioned above), they provide an algorithm with a regret scaling with horizon as $\log^{1+\delta}(T)$ for any $\delta>0$, using longer exploration phases.

\medskip

\citet{shi2021heterogeneous} recently adapted the well-known \cucb algorithm to heterogeneous multiplayer bandits. Their \textsc{BEACON} algorithm reaches the centralized optimal performance for arbitrary correlations between the arms, even with several optimal matchings. Adapting \cucb to the multiplayer setting is possible thanks to two main ideas. First, \cucb is run in a batched version where the players pull the maximal matching (in terms of confidence bound) for $n_p$ rounds during epoch $p$. The length of the epoch $p$ here depends on the previous epochs and is thus communicated by the leader for each new epoch during the communication phase. 
The second main idea is to use \textit{adaptive differential communication}. Namely, instead of communicating their empirical means $\hat{\mu}_k^j(p)$ to the leader at the end of epoch $p$, the players only communicate the difference $\hat{\mu}_k^j(p)-\hat{\mu}_k^j(p-1)$ between their current empirical means and the ones at the end of the previous epoch. While sending $\hat{\mu}_k^j(p)$ can be done in $p$ rounds, the difference can instead be sent in expectation in $\bigO{1}$ rounds, drastically reducing the communication cost without loss of information. Using these two techniques, \textsc{BEACON} thus reaches the $\bigO{\frac{M^2K\log(T)}{\Delta}}$ regret guarantee of \cucb, up to logarithmic terms in $K$.

\subsection{No communication} \label{sec:nocomm}

The previous section showed how the collision information can be leveraged to enable communication between players. These communication schemes are yet often unadapted to the reality, for different reasons given in \cref{sec:realistic}.
Especially, while the communication cost is small in $T$, it is large in other problem parameters such as $M$, $K$ and $\frac{1}{\Delta}$. These quantities can be large in real cognitive radio networks and the communication cost of algorithms presented in \cref{sec:fullcomm} is then significant. 

Some works instead focus on which level of regret is possible without collision information at all in the homogeneous setting. In that case, no communication can happen between the players. Naturally, these works assume a pre-agreement between players, who know beforehand $M$ and are assigned different ranks in $[M]$.

\medskip

The algorithm of \citet{liu2010}, presented in \cref{sec:limited}, provides a first algorithm using no collision information. \citet{boursier2020} later reached the asymptotic regret bound $M\sum_{k>M} \frac{\mu_{(k)}-\mu_{(M)}}{\kl{\mu_{(M)}, \mu_{(k)}}}$, adapting the main phase of \wangalgo in this setting. This bound is optimal among the class of algorithms using no collision information \citep{besson2019a}.

\medskip

Despite being asymptotically optimal, this algorithm suffers a considerable regret when the suboptimality gap $\Delta$ is close to $0$. It indeed relies on the fact that if the arm rankings of the players are the same, there is no collision, while the complementary event appears an order $\frac{1}{\Delta^2}$ of rounds.

\citet{bubeck2020c} instead focus on reaching a minimax regret scaling with the horizon as $\sqrt{T \log(T)}$ without collision information. A preliminary work \citep{bubeck2020b} proposed a first geometric solution for two players and three arms, before being extended to general numbers of players and arms with combinatorial arguments.
Their algorithm avoids any collision at all with high probability, using a colored partition of $[0,1]^K$, where a color gives a matching between players and arms. Thus, the estimation $\hat{\pmb{\mu}}^j$ of all arms by a player gives a point in $[0,1]^K$ and consequently, an arm to pull for this player. 
The key of the algorithm is that for close points in $[0,1]^K$, different matchings might be assigned, but they do not overlap, \ie if players have close estimations $\hat{\pmb{\mu}}^j$ and $\hat{\pmb{\mu}}^i$, they pull different arms.
Such a coloring implies that for some regions, players might deliberately pull suboptimal arms, but at a small cost, to avoid collisions with other players.

Unfortunately, the regret of the algorithm by \citet{bubeck2020c} still suffers a dependency $MK^{11/2}$, which increases considerably with the number of channels~$K$.

\section{Towards Realistic Considerations} \label{sec:realistic}

\cref{sec:reaching} proposes algorithms reaching very good regret guarantees for different settings. Most of these algorithms are yet unrealistic, \eg a large amount of communication occurs between the players, while only a very small level of communication is possible between the players in practice. 
The fact that good theoretical algorithms are actually bad in practice emphasizes that the model of \cref{sec:model} is not well designed. In particular, it might be too simple with respect to the real problem of cognitive radio networks.

\cref{sec:nocomm} suggests that this discrepancy might be due to the fact that the number of secondary users and channels ($M$ and $K$) is actually very large, and the dependency on these terms is as significant as the dependency in $T$. This kind of question even appears in the bandits literature for a single-player (and a very large number of arms). Recent works showed that the greedy algorithm actually performs very well in this single-player setting, confirming a behavior that might be observed in some real cases \citep{bayati2020unreasonable, jedor2021greedy}. It yet remains to study such a condition in the multiplayer setting.

\medskip

This section proposes other reasons for this discrepancy and removes some simplifications from the multiplayer model, in the hope of obtaining algorithms adapted to real-world situations.
First, the stochasticity of the reward $X_k$ is questioned in \cref{sec:nonstoch} and replaced by either Markovian, abruptly changing, or adversarial rewards.
The current collision model is then relaxed in \cref{sec:diffcoll}. It instead considers a more difficult model where players only observe a decrease in reward when colliding.
\cref{sec:noncollab} considers non-collaborative players, who can be either adversarial or strategic.
\cref{sec:dynamic} finally questions the time synchronization that is assumed in most of the literature.

\subsection{Non-stochastic Rewards}\label{sec:nonstoch}

Most existing works in multiplayer bandits assume that the rewards $X_k(t)$ are stochastic, \ie they are drawn according to the same distribution at each round. This assumption might be too simple for the problem of cognitive radio networks, and other settings can instead be adapted from the bandits literature. 
It has indeed been the case for markovian rewards, abruptly changing rewards and adversarial rewards, as described in this section.

\subsubsection{Markovian Rewards.}
A first more complex model is given by Markovian rewards. In this model introduced by \citet{anantharam1987b}, the reward $X_k^j$ of arm $k$ for player~$j$ follows an irreducible, aperiodic, reversible Markov chain on a finite space. 
Given the transition probability matrix $P_k^j$, if the last observed reward of arm $k$ for player $j$ is $x$, then player $j$ will observe $x'$ on this arm for the next pull with probability $P_k^j(x,x')$.

Given the stationary distribution $p_k^j$ of the Markov chain represented by $P_k^j$, the expected reward of arm $k$ for player $j$ is then equal to
\begin{equation*}
\mu_k^j = \sum_{x \in \mathcal{X}} x p_k^j(x),
\end{equation*}
where $\mathcal{X} \subset [0, 1]$ is the state space. The regret then compares the performance of the algorithm with the reward obtained by pulling the maximal matching with respect to $\mu$ at each round.

\medskip

\citet{anantharam1987b} proposed an optimal centralized algorithm for this setting, based on a UCB strategy. 
\citet{kalathil2014} later proposed a first decentralized algorithm for this setting, following the same lines as their algorithm described in \cref{sec:fullcomm} for the stochastic case. Recall that it uses explicit communication between players to assign the arms to pull.
The only difference is that the UCB index has to be adapted to the markovian model. The uncertainty is indeed larger in this setting, and the regret is thus larger as well. 
\citet{bistritz2020b} also showed that the GoT algorithm can be directly extended to this model, with proper tuning of its different parameters.

In more recent work, \citet{gafni2021} instead consider a restless Markov chain, \ie the state of an arm changes according to the Markov chain at each round, even when it is not pulled. The Gilbert-Elliot model, which is often considered as a good model for the activity of primary users \citep{tekin2011online}, is a particular case of restless Markov chain. 
Using an ETC approach, they were able to reach a stable matching in a logarithmic time. Their result yet assumes knowledge of the suboptimality gap $\Delta$ and the uniqueness of the stable matching. 
Moreover, it only reaches a stable matching instead of an optimal one, similarly to the algorithms described in \cref{sec:limited} for the heterogeneous setting.
The main difficulty of the restless setting is that the exploration phase has to be carefully done in order to correctly estimate the expected reward of each arm. This adds a dedicated random amount of time at the start of every exploration phase.

\subsubsection{Abruptly Changing Rewards.}

Although Markovian rewards are closer to reality, the resulting algorithms are very similar to the stochastic case. Indeed, the goal is still to pull the arm with the maximal mean reward, with respect to the stationary distribution.
A stronger model assumes instead that the expected rewards abruptly change over time, \eg the mean vector $\mu$ is piecewise constant in time, and each change is a \textit{breakpoint}. 
Even in the single-player case, this problem is far from being solved \citep[see e.g.][]{auer2019, besson2020}.

\citet{wei2018} considered this setting for the homogeneous multiplayer bandits problem. Assuming a pre-agreement on the ranks of players, they propose an algorithm with a regret scaling with the horizon as $T^{\frac{1+\nu}{2}} \log(T)$, when the number of breakpoints is $\bigO{T^\nu}$. 
Players use UCB indices computed on sliding windows of length $\bigO{t^{\frac{1-\nu}{2}}}$, \ie they compute the indices using only the observations of the last $t^{\frac{1-\nu}{2}}$ rounds. Based on this, player $k$ either rotates on the top-$M$ indices or focuses on the $k$-th best index to avoid collisions with other players.

\subsubsection{Adversarial Rewards.}

The hardest model for rewards is the adversarial case, where the rewards are fixed by an adversary. In this case, the goal is to provide a minimax regret bound that holds under any problem instance. 
\citet{bubeck2020a} showed that for an \textit{adaptive} adversary, who chooses the rewards $X_k(t)$ of the next round based on the previous decisions of the players, the regret lower bound is linear with $T$. The literature thus focuses on an \textit{oblivious} adversary, who chooses beforehand the sequences of rewards $X_k(t)$.

\medskip

\citet{bande2019a} proposed a first algorithm based on the celebrated EXP.3 algorithm. The EXP.3 algorithm pulls the arm $k$ with a probability proportional to $e^{-\eta S_k}$ where $\eta$ is the learning rate and $S_k$ is an estimator of $\sum_{s < t} X_k(s)$. Not all the terms of this sum are observed, justifying the use of an estimator.
To avoid collisions, \citet{bande2019a} run EXP.3 in blocks of size $\sqrt{T}$. In each of these blocks, the players start by pulling with respect to the probability distribution of EXP.3 until finding a free arm. Afterward, the player keeps pulling this arm until the end of the block. This algorithm yields a regret scaling with time as $T^{3/4}$. Dividing EXP.3 in blocks thus degrades the regret by a factor $T^{1/4}$ here.

\citet{alatur2020} proposed a similar algorithm, with a leader-followers structure. At the beginning of each block, the leader communicates to the followers the arms they have to pull for this block, still using the probability distribution of EXP.3. Also, the size of each block is here of order $T^{1/3}$, leading to a better regret, scaling with time as $T^{2/3}$.

\citet{shi2020b} later extended this algorithm to the no-sensing setting. They introduce the \textit{attackability} of the adversary, which is the length of the longest possible sequence of $X_k=0$ on an arm. Knowing this quantity $W$, a bit can indeed be correctly sent in time $W+1$.
When the attackability is of order $T^\alpha$ and $\alpha$ is known, the algorithm of \citet{alatur2020} can then be adapted and yields a regret scaling with time as $T^{\frac{2+\alpha}{3}}$.

The problem is much harder when $\alpha$ is unknown. In this case, the players estimate $\alpha$ by starting from $0$ and increasing this quantity by $\varepsilon$ at each communication failure. To keep the players synchronized with the same estimate of $\alpha$, the followers then report the communication failure to the leader. These reports are crucial and can also fail because of $0$ rewards. \citet{shi2020b} here use error detection code and randomized communication rounds to avoid such situations.

\medskip

\citet{bubeck2020a} were the first to propose an algorithm with a regret scaling with the horizon as $\sqrt{T}$ for the collision sensing setting, but only with two players. Their algorithm works as follows: a first player follows a low switching strategy, \eg she changes the arm to pull after a random number of times of order $\sqrt{T}$, while the second player follows a high-switching strategy, given by EXP.3, on all the arms except the one pulled by the first player. At each change of arm for the first player, a communication round then occurs so that the second player is aware of the choice of the first one. 

This algorithm requires shared randomness between the players, as the first player changes her arm at random times. Yet, the players can choose a common \textit{seed} during the initialization, avoiding the need for this assumption.

\medskip

\citet{bubeck2020a} also proposed an algorithm with a regret scaling with time as $T^{1-\frac{1}{2M}}$ for the no-sensing setting. For two players, the first, low-switching player runs an algorithm on the arms $\lbrace 2, \ldots, K\rbrace$ and divides the time into fixed blocks whose length is of order $\sqrt{T}$. Meanwhile on each block, the high-switching player runs EXP.3 on an increasing set $S_t$ starting from $S_t = \lbrace 1 \rbrace$. At random times, this player pulls arms outside $S_t$ and adds them to the set $S_t$ if they get a positive reward. The arm pulled by the first player is then never added to $S_t$.
For more than two players, \citet{bubeck2020a} generalize this algorithm using blocks of different sizes for different players.

\subsection{Different Collision Models} \label{sec:diffcoll}

As shown in \cref{sec:fullcomm}, the collision information allows communication between the different players.
The discrepancy between the theoretical and practical algorithms might be due to the collision model, which assumes that a collision systematically corresponds to a $0$ reward. Such a model is motivated by the carrier-sense multiple access with collision avoidance protocol, which is for instance used in WiFi.

\medskip

\textbf{Non-zero collision reward.}

In some modern wireless systems, central scheduling may allow to share the resource between users, \eg using multiplexing \citep{sesia2011lte}. Reduced-rate communications can then be successful in the presence of multiple users. 
In that case, collisions only lead to a decrease in reward and not necessarily a $0$ reward. 
Also, the number of secondary users can exceed the number of channels. 
This harder setting was introduced by \citet{tekin2012}. In the heterogeneous setting, when player~$j$ pulls an arm~$k$, the expectation of the random variable $X_k^j(t)$ also depends on the total number of players pulling this arm. The problem parameters are then given by the functions $\mu_k^j(m)$ which give the expectation of $X_k^j$ when exactly $m$ players are pulling the arm $k$.
Naturally, the function $\mu_k^j$ is non-increasing in~$m$. The regret then compares the cumulative reward with the one obtained by the best allocation of players through the different arms.

\citet{tekin2012} proposed a first ETC algorithm when players know the suboptimality gap of the problem and always observe the number of players pulling the same arm as they do.
These assumptions are pretty strong and are not considered in the more recent literature. 

\citet{bande2019a} also proposed an ETC algorithm, still with the prior knowledge of the suboptimality gap. During the exploration, players pull all arms at random.
The main difficulty is that when players observe a reward, they do not know how many other players are also pulling this arm. 
\citet{bande2019a} overcome this problem by assuming that the decrease in mean reward when an additional player pulls an arm is large enough with respect to the noise. As a consequence, the observed rewards on a single arm can then be perfectly clustered and each cluster exactly corresponds to the observations for a given number of players pulling the arm. 

\medskip

In practice, this assumption is actually very strong and means that the observed rewards are almost noiseless. \citet{magesh2019b} instead assume that all the players have different ranks. 
Thanks to this, they can coordinate their exploration, so that all players can explore each arm $k$ with a known and fixed number of players $m$ pulling it. Exploring for all arms and all numbers of players $m$ then allows the players to know their own expectations $\mu^j_k(m)$ for any $k$ and $m$.
From there, the players can reach the optimal allocation using a Game of Thrones routine similar to Algorithm~\ref{alg:GoT}. This work thus extended the results of \citet{bistritz2020b} to the harder setting of non-zero rewards in case of collision.
\citet{bande2021} recently used a similar exploration for the homogeneous setting. 

\medskip

When the arm mean is exactly inversely proportional to the number of players pulling it, \ie $\mu_k^j(m) = \frac{\mu_k^j(1)}{m}$, \citet{boyarski2021distributed} exploit this assumption to design a simple algorithm with a regret scaling with time as $\log^{3+\delta}(T)$. 
During the exploration phase, all players first pull each arm $k$ altogether and estimate $\mu_k^j(M)$. From there, they add a block where they pull the arm $1$ with probability $\frac{1}{2}$, allowing to estimate $M$ and thus the whole functions $\mu_k^j$. The optimal matching is then assigned following a GoT subroutine.


\medskip

\textbf{Competing bandits.}\label{sec:comp_bandits1}

A recent stream of literature considers another collision model where only one of the pulling players gets the arm reward, based on the preferences of the arm. This setting, introduced by \citet{liu2019}, is discussed in \cref{sec:comp_bandits2}.

\subsection{Non-collaborative Players}\label{sec:noncollab}

Assuming perfectly collaborative players might be another oversimplification in multiplayer bandits. A short survey by \citet{attar2012} presents the different security challenges for cognitive radio networks. Roughly, these threats are divided into two types: \textit{jamming attacks} and \textit{selfish players}. These security threats thus appear as soon as players are no more fully cooperative.

\medskip

\textbf{Jammers.}
Jamming attacks can happen either from agents external to the network or directly within the network. Their goal is to deteriorate the performance of other agents as much as possible.
In the first case, it can be seen as malicious manipulations of the rewards generated on each arm. \citet{wang2015} then propose to consider the problem as an adversarial instance and use EXP.3 algorithm in the centralized setting.

\citet{sawant2019} on the other side consider jammers directly within the network. The jammers thus aim at causing a maximal loss to the other players by either pulling the best arms or creating collisions. Without any restriction on the jammers' strategy, they can perfectly adapt to the other players' strategy and incur tremendous losses. Because of this, they restrict the jammers' strategy to pulling at random the top $J$-arms for any $J \in [K]$, either in a centralized (no collision between jammers) or decentralized way. The players then use an ETC algorithm, where the exploration aims at estimating the arm means, but also both the number of players and the number of jammers. Afterward, they exploit by sequentially pulling the top $J$-arms where $J$ is chosen to maximize the earned reward. 

\medskip

\textbf{Fairness.}
A first attempt at preventing selfish behaviors is to ensure \textit{fairness} of the algorithms, as noted by \citet{attar2012}. A fair algorithm should not favor some player with respect to another. 
In the homogeneous setting, a first definition of fairness is to guarantee the same expected rewards to all players \citep{besson2018}. 
Note that all symmetric algorithms (\ie no prior ranking of the players) ensure this property. A stronger notion would be to guarantee the same asymptotic rewards to all players without expectation,\footnote{This notion is defined \textit{ex post}, as opposed to the previous one which is \textit{ex ante}.} which can still be easily reached by making the players sequentially pull all the top-$M$ arms in a round robin fashion when exploiting.

\medskip

The notion of fairness becomes intricate in the heterogeneous setting, since it can be antagonistic to the maximization of the collective reward.
\citet{bistritz2021one} consider max-min fairness, which is broadly used in the resource allocation literature. Instead of maximizing the sum of players' rewards, the goal is to maximize the minimal reward earned by any player at each round.
They propose an ETC algorithm that determines the largest possible $\gamma$ such that all players can earn at least $\gamma$ at each round. For the allocation, the players follow a specific Markov chain to determine whether players can all reach some given~$\gamma$.
If instead, the objective is for each player $j$ to earn at least $\gamma_j$ for some known and feasible vector $\pmb{\gamma}$, there is no need to explore which is the largest possible $\gamma$ and the regret becomes constant.

\medskip

\textbf{Selfish players.}

While jammers try to incur a huge loss to other players at any cost, selfish players have a different objective: they maximize their own individual rewards. In the algorithms mentioned so far, a selfish player could largely improve her earned regret at the expense of the other players. \citet{boursier2020} propose algorithms robust to selfish players, being $\varepsilon$-Nash equilibria, where $\varepsilon$ scales logarithmically with time. Without collision information, they adapt \wangalgo without communication between the players. The main difficulty comes from designing a robust initialization protocol to assign ranks and estimate $M$.
With collision information, they even show that robust communication-based algorithms are possible, thanks to a Grim Trigger strategy that punishes all players as soon as a deviation from the collective strategy is detected. The centralized performances are thus still possible with selfish players.

Reaching the optimal matching might not be possible in the heterogeneous case because of the strategic feature of the players. Instead, they focus on reaching the average reward when following the Random Serial Dictatorship algorithm, which has good strategic guarantees in this setting \citep{abdulkadirouglu1998random}. 

\medskip

\citet{xu2023competing} later proposed an algorithm that is robust to selfish behaviors with non-zero reward collisions. More precisely, they consider a homogeneous setting where the reward of an arm is split (unevenly) between the players pulling it. In this model, reaching maximal social welfare might be impossible with selfish players (price of anarchy larger than $1$). Instead, they propose an algorithm inspired by \wangalgo, which aims at pulling the unique (up to players' permutation) Nash equilibrium of the game. It then yields an asymptotically optimal regret with respect to the utility of this Nash equilibrium.

\medskip

\citet{branzei2019} consider a different strategic multiplayer bandits game. First, their model is collisionless and players still earn some reward when pulling the same arm. Also, they consider two players and a one-armed Bayesian bandit game. Players observe both their obtained reward and the choice of the other player.
In this setting, they compare the different Nash equilibria when players are either collaborative (maximizing the sum of two rewards), neutral (maximizing their sole reward), and competitive (maximizing the difference between their reward and the other player's reward).
Players tend to explore more when cooperative and less when competitive. A similar behavior is intuitive in the classical model of multiplayer bandits as selfish players would more aggressively commit on the best arms to keep them for a long time.

\subsection{Beyond Time Synchronization}

Most of the multiplayer algorithms depend on a high level of synchronization between the players. In particular, they assume two major simplifications: the players are assumed to be synchronous and also start and end the game at the same time.

\medskip

\textbf{Asynchronous players.} 
First, the synchronous assumption considers that all players pull arms simultaneously (we say they are active) at each time step. 
Although assuming a common time discretization might be reasonable, \eg through the use of \textit{orthogonal frequency-division multiplexing} \citep{schmidl1997robust}, the devices might not be active at each timestep in practice. 
For example in Europe, the 868 MHz bandwidth, which is used for IoT transmissions, limits the duty cycle at $1\%$ to avoid saturation of the network. This means that each device can emit at most $1\%$ of the time. As a consequence, each device is only active a small fraction of the time, which is specific to each device.

\medskip

\citet{bonnefoi2017} first considered the \textit{asynchronous} setting, where each player is active with a probability $p$ at each time step. They first focus on offline policies, as even determining the optimal policy is not straightforward here, and compare empirically single-player online strategies (UCB and TS) to these baselines.

While considered homogeneous in this work, the activation rates might be heterogeneous in practice, for example, due to the heterogeneity of the devices and their applications in IoT networks. \citet{dakdouk2021collaborative} consider activation probabilities $p_m$ varying among the players. Computing the optimal policy becomes even more intricate in this case, and they propose an offline greedy policy, which reduces to the optimal policy with homogeneous activation rates. They also propose a greedy policy with fairness guarantees.
Two Explore-then-Commit algorithms based on these baseline policies are then proposed and yield regret scaling with time as $T^{2/3}$.

\citet{richard2023constant} recently proposed two centralized algorithms in the presence of homogeneous activation rates.  First, a UCB-based algorithm yields a minimax regret scaling with the horizon as $\sqrt{T\log(T)}$. Second, they propose Cautious Greedy algorithm, which greedily pulls the best possible assignment, but only assigns no player to an arm if it is confident enough that the optimal assignment does not assign any player to this arm. This algorithm has two different regimes of regret: in the case where the optimal policy assign at least one player to each arm, it yields a regret constant in time; while it is scales logarithmically with time when some arm is assigned no player by the optimal policy. Extending this algorithm to a decentralized setting yet remains open and an interesting direction for future work.

\medskip

\textbf{Dynamic model.}\label{sec:dynamic}

Second, the players are assumed to respectively start and end the game at times $t=1$ and $T$. In practice, secondary users enter and leave the network at different time steps, making this assumption oversimplifying.
The dynamic model removes this assumption: players enter and leave the bandits instance at different times. 

The MEGA algorithm of \citet{avner2014} was the first proposed algorithm to deal with this dynamic model. The exact same algorithm as the one described in \cref{sec:limited} still reaches a regret of order $NT^{\frac{2}{3}}$ in this case when omitting the dependency in $K$ and the means $\mu$. $N$ is here the total number of players entering or leaving the network.

\medskip

In general, $N$ is assumed to be sublinear in $T$ as otherwise players would enter and leave the network too quickly to learn the different problem parameters. \citet{rosenski2016} propose to divide the game duration into $\sqrt{NT}$ epochs of equal size and run independently the \musical algorithm on each epoch. The number of failing epochs is at most $N$ and their total incurred regret is thus of order $\sqrt{NT}$. Precisely, the total incurred regret by this algorithm is $\bigO{\sqrt{NT}\frac{K^2 \log(T)}{\Delta^2}}$.

This technique can be used to adapt any static algorithm but requires the knowledge of the number of entering/leaving players $N$, as well as a shared clock between players, to remain synchronized on each epoch. Because it also works in time windows of size $\sqrt{T}$, the algorithm of \citet{bande2019a} in the adversarial setting still has a regret scaling with time in $T^{\frac{3}{4}}$ in the dynamic setting.

On the other hand, \citet{bande2019a, bande2021} propose to adapt their static algorithms, with epochs of linearly increasing size. Players do not need to know $N$ here, but instead need a stronger shared clock, since they also need to know in which epoch they currently are.

\medskip

Besides requiring some strong assumption on either players' knowledge or synchronization, this kind of technique also leads to large dependencies in $T$. Players indeed run independent algorithms on a large number of time windows and thus suffer a considerable loss when summing over all the epochs.

To avoid this kind of behavior, \citet{boursier2018} consider a simpler dynamic setting, where players can enter at any time but all leave the game at time $T$.
They propose a no-sensing ETC algorithm, which requires no prior knowledge and no further assumption. The idea is that exploring uniformly at random is robust to the entering/committing of other players. The players then try to commit to the best-known available arm. This algorithm leads to a regret $\bigO{\frac{NK\log(T)}{\bar{\Delta}^2}}$, with $\bar{\Delta}_{(M)}=\min_{i\in[M]} \mu_{(i)}-\mu_{(i+1)}$.

On the other hand, the algorithm by \citet{darak2019} recovers from the event of entry/leave of a player after some time depending on the problem parameters. 
However, if enter/leave events happen in a short time window, the algorithm has no guarantees. This algorithm is thus adapted to another simpler dynamic setting, where the events of entering or leaving of a new player are separated from a minimal duration.

\section{Related Problems}\label{sec:alternateproblems}

This section introduces related problems that have also been considered in the literature. All these models consider a bandits game with multiple agents and some level of interaction between these agents. Because of the similarities with the multiplayer bandits problem considered in this survey, the methods and techniques mentioned above can be directly used or adapted to these related problems.

The widely studied problem of multi-agent bandits is first mentioned. \cref{sec:comp_bandits2} then introduces the problem of competing bandits, motivated by matching markets. \cref{sec:queuing} finally discusses the problem of queuing systems, motivated by packet routing in servers.

Although multiplayer bandits is a particular instance of Multi-Agent Reinforcement Learning (MARL), MARL is not discussed in this survey. MARL is indeed a much more complicated problem, since determining the optimal policy (with perfect knowledge of problem parameters) is a hard problem on its own. On the other hand, determining the optimal policy in multiplayer bandits is straightforward once the arm means are known. This difference in difficulties comes from the simple structure of the rewards in multiplayer bandits, and especially because of its simple dependency in other players' actions. 
In particular, designing centralized MARL algorithms is still of major interest in the current MARL literature. Moreover because of the complexity of the problem, MARL constitutes a research field on its own and we refer the reader to \citet{busoniu2008comprehensive,zhang2021multi} for dedicated surveys.

\subsection{Multi-agent Bandits}  \label{sec:multiagent}

The multi-agent bandits problem (also called cooperative bandits or distributed bandits) introduced by \citet{awerbuch2008} considers a bandit game played by $M$ players. 
Motivated by distributed networks where agents can share their cumulated information, players here encounter no collision when pulling the same arm: their goal is to collectively determine the best arm.
While running a single-player algorithm such as UCB already yields regret guarantees, players can improve their performance by collectively sharing some information. The way players can communicate yet remains limited: they can only directly communicate with their neighbors in a given graph $\mathcal{G}$.

\medskip

This problem has been widely studied in the past years, and we do not claim to provide an extensive review of its literature.
Many algorithms are based on a gossip procedure, which is widely used in the more general field of decentralized computation. Roughly, a player $i$ updates its estimates $\hat{x}^i$ by averaging (potentially with different weights) with the estimates $\hat{x}^j$ of her neighbors $j$. Mathematically, the estimated vector $\pmb{\hat{x}}$ is updated as follows:
\begin{equation*}
\pmb{\hat{x}} \gets P \pmb{\hat{x}}, 
\end{equation*}
where $P$ is a communication (bistochastic) matrix. To respect the communication graph structure, $P_{i,j}>0$ if and only if the edge $(i,j)$ is in the graph $\mathcal{G}$. $P$ thus gives the weights used to average these estimates.

\citet{szorenyi2013} propose an $\varepsilon$-greedy strategy with gossip-based updates, while \citet{landgren2016} propose a gossip UCB algorithm. Their regret decomposes in two terms: a centralized term approaching the regret incurred by a centralized algorithm and a second term, which is constant in $T$ but depends on the spectral gap of the communication matrix $P$ and can be seen as the \textit{delay} to pass a message through the graph with the gossip procedure.
Improving this graph-dependent term is thus the main focus of many works. \citet{martinez2018} propose a UCB algorithm with gossip acceleration techniques, improving upon previous work \citep{landgren2016}.

\medskip

Another common procedure is to elect a leader in the graph, who sends the arm (or distribution) to pull to the other players. In particular, \citet{wang2020} adapt the \wangalgo algorithm described in \cref{sec:fullcomm} to the multi-agent bandits problem. The leader is the only exploring player and sends her best empirical arm to the other players. Despite having an optimal regret bound in $T$, the second term of the regret due to communication scales with the diameter of the graph $\mathcal{G}$. This algorithm only requires for the players to send $1$-bit messages at each time step, while most multi-agent bandits works assume that the players can send real messages with infinite precision.

In the adversarial setting, \citet{bar2019} propose to elect \textit{local} leaders who send a play distribution, based on EXP.3, to their followers. Instead of focusing on the collective regret as usually done, they provide good individual regret guarantees, leading to a fair algorithm.

\medskip

Another line of work assumes that a player observes the rewards of all her neighbors at each time step. \citet{cesa2019a} even assume to observe rewards of all players at distance at most $d$, with a delay depending on the distance of the player. EXP.3 with smartly chosen weights then allows reaching a small regret in the adversarial setting.

More recent works even assume that the players are asynchronous, \ie players are active at a given time step with some activation probability. This is for example similar to the model of \citet{bonnefoi2017} in the multiplayer setting. \citet{cesa2020} then use an Online Mirror Descent based algorithm for the adversarial setting. \citet{della2021} extended this idea to the combinatorial setting, where players can pull multiple arms.

\medskip

Similarly to multiplayer bandits, the problem of multi-agent bandits is wide and many directions remain to be explored. For instance, \citet{vial2020} recently proposed an algorithm that is robust to malicious players. While malicious players cannot create collisions on purpose here, they can send corrupted information to their neighbors, leading to bad behaviors.

\subsection{Competing Bandits}
 \label{sec:comp_bandits2}

The problem of competing bandits was first introduced by \citet{liu2019}, motivated by decentralized learning processes in matching markets. This model is very similar to the heterogeneous multiplayer bandits: they only differ in their collision model. Here, arms also have preferences over players: $j \succ_k j'$ means that the arm $k$ prefers being pulled by the player $j$ over $j'$. When several players pull the same arm $k$, only the top-ranked player for arm $k$ gets its reward, while the others receive no reward. Mathematically the collision indicator is here:
\begin{gather*}
\coll^j_{k}(t) = \one{\exists j' \succ_k j \text{ such that } \pull^{j'}(t) = k}.
\end{gather*}
As often in bipartite matching problems, stable matchings constitute the oracle baselines. A matching is \textit{stable} if any unmatched pair $(j,k)$ would prefer to be matched. Mathematically, this corresponds to the following definition.
\begin{definition}
A matching $\pi: [M] \to [K]$ is stable if for any $j \neq j'$, either $\mu^j_{\pi(j)} > \mu^j_{\pi(j')}$ or $j' \succ_{\pi(j')} j$ and for any unmatched arm $k$, $\mu^j_{\pi(j)}>\mu^j_k$. 
\end{definition}

Several stable matchings can exist. 
Two different definitions of individual regret then appear. First the \textit{optimal regret} compares with the best possible arm for player $j$ in a stable matching, noted $\bar{k}_j$:
\begin{equation*}
\bar{\reg}_j(T) = \mu^j_{\bar{k}_j} T - \sum_{t=1}^T \mu_{\pull^j(t)}^j \cdot (1-\coll^j_{\pull^j(t)}(t)).
\end{equation*}
Similarly, the pessimal regret is defined with respect to the worst possible arm for player $j$ in a stable matching, noted $\munderbar{k}_j$:
\begin{equation*}
\munderbar{\reg}_j(T) = \mu^j_{\munderbar{k}_j} T - \sum_{t=1}^T \mu_{\pull^j(t)}^j \cdot (1-\coll^j_{\pull^j(t)}(t)).
\end{equation*}

\citet{liu2019} propose a centralized UCB algorithm, where at each time step, the players send their UCB indexes to a central agent. This agent computes the optimal stable matching based on these indexes using the celebrated Gale Shapley algorithm and the players then pull according to the output of Gale Shapley algorithm. Although being natural, this algorithm only reaches a logarithmic regret for the pessimal definition, but can still incur a linear optimal regret.

\citet{cen2021} showed that a logarithmic optimal regret is reachable for this algorithm if the platform can also choose transfers between the players and arms, which here play a symmetric role. The idea is to smartly choose the transfers so that the optimal matching in social welfare is the only stable matching when taking into account these transfers. Their notion of equilibrium is yet weak, as the players and arms do not negotiate the transfers fixed by the platform.
\citet{jagadeesan2021learning} instead consider the stronger notion of equilibrium where the agents also negotiate these transfers. They define the \textit{subset instability}, which represents the distance of a market outcome from equilibrium and is considered as the incurred loss at each round. Using classical UCB-based algorithms, they are then able to minimize the regret with respect to this measure of utility loss.

\medskip

\citet{liu2019} propose an ETC algorithm reaching a logarithmic optimal regret without any transfer. After the exploration, the central agent computes the Gale Shapley matching which is pulled until $T$. A decentralized version of this algorithm is even possible, as Gale Shapley can be run in times $N^2$ in a decentralized way when observing the collision indicators $\coll^j_k$. This decentralized algorithm yet requires prior knowledge of $\Delta$. \citet{basu2021} extend this algorithm without knowing $\Delta$, but the optimal regret incurred then scales with time as $\log^{1+\varepsilon}(T)$ for any parameter $\varepsilon>0$.

\medskip

\citet{liu2020} also propose a decentralized UCB algorithm with a collision avoidance mechanism. Yet their algorithm requires for the players to observe the actions of all other players at each time step and incurs a pessimal regret of order $\log^2(T)$ with exponential dependence in the number of players.

\medskip

Because of the difficulty of the general problem, even with collision sensing, another line of work focuses on simple instances of arm preferences. For example, when players are \textit{globally ranked}, \ie all the arms have the same preference orders $\succ_k$, there is a unique stable matching. Moreover, it can be computed with the Serial Dictatorship algorithm, where the first player chooses her best arm, the second player chooses her best available arm, and so on.
In particular, the algorithm of \citet{liu2020} yields a logarithmic regret without exponential dependency in this case.

Using this simplified structure, \citet{sankararaman2020} also propose a decentralized UCB algorithm with a collision avoidance mechanism. Working in epochs of increasing size, players mark as blocked the arms declared by players of smaller ranks and only play UCB on the unblocked arms. Their algorithm yields a regret bound close to the lower bound, which is shown to be at least of order $R_j(T)=\Omega\left(\max\left(\frac{(j-1)\log(T)}{\Delta^2}, \frac{K\log(T)}{\Delta}\right)\right)$ for some instance.\footnote{Optimal and pessimal regret coincide here as there is a unique stable matching.} The first term in the $\max$ is the number of collisions encountered with players of smaller ranks, while the second term is the usual regret in single-player stochastic bandits.

Serial Dictatorship can lead to unique stable matching even in more general settings than globally ranked players. In particular, this is the case when the preferences profile satisfies uniqueness consistency. \citet{basu2021} then adapt the aforementioned algorithm to this setting, by using a more subtle collision avoidance mechanism. 

\medskip

Recently, \citet{zhang2022matching} proposed an ETC algorithm with a communication protocol based on \sicmmab[.] The communication protocol is yet more technical as some players might not be able to communicate with each other. The arms preferences indeed fix the possible communications between player pairs. However, if two players cannot communicate, this means that one of them is always preferred over the other one; her optimal assigned arm thus does not depend on the other player's preferences. \citet{zhang2022matching} thus propose a \textit{multi-layered communication} that still allows to exchange enough information between the players to determine the optimal matching. 
Their proposed algorithm allows to reach a collective optimal regret scaling as $\sum_{j=1}^M \bar{\reg}_j(T) = \bigO{\frac{MK}{\Delta^2}\log(T)}$, which is thus the first $\log(T)$ optimal regret bound in the general decentralized setting. The $\frac{1}{\Delta^2}$ dependency yet remains sub-optimal. Improving upon this dependency and getting a sublinear minimax regret are possible directions for future work.


\subsection{Queuing Systems}\label{sec:queuing}

\citet{gaitonde2020a} extended the queuing systems introduced by \citet{krishnasamy2016} to the multi-agent setting. This problem remains largely open as it is quite recent, but similarly to the competing bandits problem, it might benefit from multiplayer bandits approaches.

In this model, players are queues with arrival rates $\lambda_i$. At each time step, a packet is generated within the queue $i$ with probability $\lambda_i$, and the arm (server)~$k$ has a clearing probability $\mu_k$. 
This model assumes some asynchronicity between the players as they have different arrival rates $\lambda_i$. Yet it remains different from the usual asynchronous setting \citep{bonnefoi2017}, as players can play as long as they have remaining packets.

When several players send packets to the same arm, only the oldest packet is treated and is then cleared with probability $\mu_k$; \ie when colliding, only the queue with the oldest packet gets to pull the arm.
A queue is said \textit{stable} when its number of packets grows almost surely as $\smallo{t^\alpha}$ for any $\alpha>0$.

\medskip

A crucial quantity of interest is the largest $\eta \in \mathbb{R}$ such that
\begin{equation*}
\eta\sum_{i=1}^k \lambda_{(i)} \leq \sum_{i=1}^k \mu_{(i)} \qquad \text{ for any } k \in [M].
\end{equation*}
In the centralized case, stability of all queues is possible if and only if $\eta > 1$.
\citet{gaitonde2020a} study whether a similar result is possible, even in the decentralized case where players are strategic. They first show that if players follow \textit{suitable} no-regret strategies, stability is reached if $\eta>2$. Yet, for smaller values of $\eta$, no regret strategies can still lead to unstable queues.

\medskip

In a subsequent work \citep{gaitonde2020b}, they claim that minimizing the regret is not a good objective as it leads to myopic behaviors of the players. Players here might prefer to be patient, as there is a carryover effect over the rounds. The issue of a round indeed depends on the past since a server treats the oldest packet sent by a player. A player thus can have an interest in letting the other players to clear their packets, as it guarantees her to avoid colliding with them in the future.

To illustrate this point, \citet{gaitonde2020b} consider the following game: all players have perfect knowledge of $\lambda$ and $\mu$ and play repeatedly a fixed probability distribution $\pmb{p}$. The cost incurred by a player is then the asymptotic value $\lim_{t\to+\infty} \frac{Q^i_t}{t}$, where $Q^i_t$ is the age of the oldest remaining packet of player~$i$ at time $t$. In this game, strategic players are even stable in situations where $\eta \leq 2$ as claimed by \cref{thm:queuingnash} below.
\begin{theorem}[\citealt{gaitonde2020b}]\label{thm:queuingnash}
If $\eta>\frac{e}{e-1}$ and all players follow a Nash equilibrium of the game described above, the system is stable.
\end{theorem}
The limit ratio $\frac{e}{e-1}$ thus corresponds to the \textit{price of anarchy} of this game. Yet this result holds only with prior knowledge of the game parameters and the price of ``learning'' remains unknown. 
The policy regret \citep{arora2012online} can be seen as a patient version of regret here, as it takes into account the long-term effects of queues' actions on future rewards. \citet{sentenac2021decentralized} show that no policy-regret strategies are no better than no regret strategies in general. In particular, they show that no policy-regret strategies can still be unstable as soon as $\eta<2$. 

They instead propose a particular decentralized strategy, such that the system is stable for any $\eta >1$ when all the queues follow this strategy, thus being comparable to centralized performances. Similar to many multiplayer bandits algorithms, this strategy takes advantage of synchronization between the players, and extending this kind of result to the dynamic/asynchronous setting remains challenging.

\medskip

\citet{freund2022efficient} considered a different preferences' model by the servers, which allows the players to give a bidding amount when pulling an arm. The player with the highest bid then gets to pull the arm. This allows to design a simpler decentralized algorithm, based on ascending-price auction. Besides yielding better finite time bounds on the number of left packets with respect to \citet{sentenac2021decentralized}, their algorithm is valid in the more general no-sensing, heterogeneous setting.

\section{Conclusion}

The multiplayer bandits problem has been widely studied in the past years. In particular, centralized optimal regret bounds are reachable in common models, by leveraging collision information as a way of communicating between players. However, these optimal algorithms might seem \textit{ad hoc} and are undesirable in real cognitive radio networks.
This suggests that the current formulation of the multiplayer bandits problem is oversimplified and does not reflect real situations and leads to undesirable optimal algorithms.
Several more realistic models were suggested in the literature to overcome this issue. In particular, the dynamic and asynchronous models seem of crucial interest since time synchronization is not verified in practice and seems crucial to communicate through collision information between players.
We personally believe that developing efficient strategies for both asynchronous and dynamic settings is a major direction for future research, which should lead to implementable algorithms in real applications. 
More generally, studying further non-stationary environments (\eg in arm means or number of players) in multiplayer bandits is of crucial interest. 
Some preliminary works focus on these aspects, but yet remain far from solving the general settings.

Besides theoretical investigation of multiplayer bandits, successfully implementing multiplayer bandits algorithms in real world applications such as IoT networks remains an obvious direction of interest, that could lead to a much better quality of service of real world networks and would boost the interest in multiplayer bandits.

\medskip

Besides its main application for communication networks, the multiplayer bandits problem is also related to several sequential multi-agent problems, notably motivated by distributed networks, matching markets and packet routing. Exploring further the potential relations that might exist between these different problems could be of great interest to any of them.
Although the problem of multiplayer bandits seems ``solved'' for its classical formulation, proposing efficient algorithms adapted to real applications remains largely open.

%

\newpage
\appendix

\section{Classical Stochastic Multi-Armed Bandits}\label{sec:banditsintro}

This section shortly describes the stochastic MAB problem, as well as the main results and algorithms for this classical instance, which provide insights for the different algorithms and results in the multiplayer bandits literature. We refer the reader to \citep{bubeck2012, lattimore2018bandit, slivkins2019introduction} for extensive surveys on MAB and for the adversarial setting.


\subsection{Model and Lower Bounds}

At each time step $t \in [T]$, the agent pulls an arm $\pi(t) \in [K]$ among a finite set of actions, where $T$ is the game horizon. When pulling the arm $k$, she observes and receives the reward $X_k(t)$ of mean $\mu_k = \mathbb{E}[X_k(t)]$. This observation $X_k(t)$ is then used by the agent to choose the arm to pull in the next rounds.
The random variables $(X_k(t))_{t=1, \ldots, T}$ are independent, identically distributed and bounded in $[0,1]$ in the following.

The goal of the agent is to maximize her cumulated reward. Equivalently, she aims at minimizing her regret, which is the difference between the maximal expected reward of an agent knowing beforehand the arms' distributions and the actual earned reward until the game horizon $T$. It is formally defined as
\begin{equation*}
R(T) = T\mu_{(1)} - \mathbb{E}\left[\sum_{t=1}^T \mu_{\pi(t)}\right],
\end{equation*}
where the expectation holds over the actions $\pi(t)$ of the agent and $\mu_{(k)}$ denotes the $k$-th largest mean. 

The player only observes the reward $X_k(t)$ of the pulled arm and not those associated to the non-pulled arms. Because of this bandit feedback, the player must balance between \textbf{exploration}, \ie estimating the arm means by pulling all arms sufficiently, and \textbf{exploitation}, by pulling the seemingly optimal arm. 
%
\begin{definition}
We call an algorithm uniformly good for if for every parameter configuration $\pmb{\mu}, K$ and $\alpha>0$, it yields $R(T) = \smallo{T^\alpha}$.
\end{definition}
The cumulated reward is of order $T\mu_{(1)}$ for an asymptotically consistent algorithm. The regret is instead a more refined choice of measure, since it captures the second order term of the cumulated reward in this case. 
First, \cref{introthm:lower} lower bounds the achievable regret in the classical stochastic MAB. We call an instance of stochastic MAB a setting with fixed $K$ and arm means $(\mu_k)_{k\in[K]}$.

\begin{theorem}[\citealt{lai1985}]\label{introthm:lower}
Consider a bandit instance with Bernoulli distributions, \ie $X_k(t)\sim\mathrm{Bernoulli}(\mu_k)$, then any uniformly good algorithm has an asymptotic regret bounded as
\begin{equation*}
\liminf_{T\to \infty} \frac{R(T)}{\log(T)} \geq \sum_{k: \mu_k < \mu_{(1)}} \frac{\mu_{(1)}-\mu_k}{\kl{\mu_{(1)}, \mu_k}},
\end{equation*}
where $\kl{p,q} = p\log\left(\frac{p}{q}\right) + (1-p)\log\left( \frac{1-p}{1-q}\right)$.
\end{theorem}
%
However, the above equation does not directly imply that the maximal regret incurred at some given time T over all the  possible instances is not linear in T (if $\mu_k$ is arbitrarily close to $\mu_{(1)}$, the right-hand side term can be arbitrarily large). This corresponds to the worst case, where the considered instance is the worst for the fixed, finite horizon $T$.
When specifying this quantity, we instead refer to the minimax regret, which is lower bounded as follows.
\begin{theorem}[\citealt{auer1995gambling}]
For any MAB algorithm and horizon $T\in\mathbb{N}$, there exists a problem instance such that
\begin{equation*}
R(T) \geq \frac{\sqrt{KT}}{20}.
\end{equation*}
\end{theorem}

\subsection{Common Algorithms}

This section describes the following bandit algorithms: $\varepsilon$-greedy, Upper Confidence Bound (UCB), Thompson Sampling and Explore-then-commit (ETC). 
%
%
The following notations are used in the remaining of this section
\begin{itemize}
\item $N_k(t) = \sum_{s=1}^{t-1} \one{\pi(s)=k}$ is the number of pulls on arm $k$ until time $t$;
\item $\hat{\mu}_k(t) = \frac{\sum_{s=1}^{t-1} \one{\pi(s)=k} X_k(t)}{N_k(t)}$ is the empirical mean of arm $k$ before time $t$;
\item $\Delta = \min \lbrace \mu_{(1)}-\mu_k >0 \mid k \in [K] \rbrace$ is the suboptimality gap and represents the hardness of learning the problem.\footnote{Note that it coincides with the definition given in \cref{sec:model}.}
\end{itemize}

\subsubsection{\texorpdfstring{$\varepsilon$}{ε}-greedy Algorithm}

\begin{algorithm}[h]
   \DontPrintSemicolon
\SetKwInOut{Input}{input}
\Input{$(\varepsilon_t)_t \in [0,1]^{\mathbb{N}}$}
\caption{\label{algo:epsgreedy}$\varepsilon$-greedy algorithm}
\lFor{$t=1, \ldots, K$}{pull arm $t$}
\vspace{0.5em}
\lFor{$t=K+1, \ldots, T$}{
$\begin{cases}
\text{pull } k \sim \mathcal{U}([K]) \text{ with probability } \varepsilon_t; \\
\text{pull } k \in \argmax_{i \in [K]} \hat{\mu}_i(t) \text{ otherwise}.
\end{cases}$
}
		\end{algorithm}

The $\varepsilon$-greedy algorithm described in Algorithm~\ref{algo:epsgreedy} is defined by a sequence $(\varepsilon_t)_{t} \in [0,1]^{\mathbb{N}}$. 
Each arm is first pulled once. Then at each round $t$, the algorithm explores with probability $\varepsilon_t$, meaning it pulls an arm chosen uniformly at random. Otherwise, it exploits, \ie it pulls the best empirical arm.
When $\varepsilon_t = 0$ for all $t$, it is called the greedy algorithm, as it always greedily pulls the best empirical arm. The greedy algorithm generally incurs a linear regret in $T$, as the best arm can be underestimated after its first pull and never be pulled again.
Appropriately choosing the sequence $(\varepsilon_t)$ instead leads to a sublinear regret, as given by \cref{thm:epsgreedy}.
\begin{theorem}[\citealt{slivkins2019introduction}]\label{thm:epsgreedy}
$\varepsilon$-greedy algorithm with exploration probabilities $\varepsilon_t=\left(\frac{K\log(t)}{t}\right)^{1/3}$ has a regret bounded as
\begin{equation*}
\singleregret(T) =\bigO{ \left(K\log(T)\right)^{1/3} T^{2/3}}.
\end{equation*}
\end{theorem}
%

\subsubsection{Upper Confidence Bound Algorithm}

\begin{algorithm}[ht]
   \DontPrintSemicolon
\SetKwInOut{Input}{input}
\lFor{$t=1, \ldots, K$}{pull arm $t$}
\lFor{$t=K+1, \ldots, T$}{
pull $k \in \argmax_{i \in [K]} \hat{\mu}_i(t) + B_i(t)$
}
\caption{\label{algo:ucb}UCB algorithm}
		\end{algorithm}

As explained above, greedily choosing the best empirical arm leads to a considerable regret, because of an under-estimation of the optimal arm mean. The UCB algorithm, given by Algorithm~\ref{algo:ucb} below, instead chooses the arm $k$ maximizing $\hat{\mu}_k(t) + B_k(t)$ at each time step, where the term $B_k(t)$ is some confidence bound ensuring that the mean estimates are (slightly) positively biased with high probability. 
\cref{thm:ucbbound} bounds the regret of the UCB algorithm with its most common choice of confidence bound.
\begin{theorem}[\citealt{auer2002}]\label{thm:ucbbound}
The UCB algorithm with $B_i(t) = \sqrt{\frac{2\log(t)}{N_i(t)}}$ verifies the following instance dependent and minimax bounds
\begin{gather}
\singleregret(T) = \bigO{\sum_{k: \mu_k<\mu_{(1)}} \frac{\log(T)}{\mu_{(1)}-\mu_k}}, \label{eq:bounducb}\\
\singleregret(T) = \bigO{\sqrt{KT\log(T)}}. \nonumber
\end{gather}
\end{theorem}
The UCB algorithm has an optimal instance dependent regret, up to some constant factor, when the arm means are bounded away from $0$ and $1$. Using finer confidence bounds, an optimal instance dependent regret is actually reachable for the UCB algorithm \citep{garivier2011}. 

\subsubsection{Thompson Sampling Algorithm}

\begin{algorithm}[ht]
   \DontPrintSemicolon
$\pmb{p} = \otimes_{k=1}^K \mathcal{U}([0,1])$ \tcp*{Uniform prior}
\For{$t=1, \ldots, T$}{
Sample $\btheta \sim \pmb{p}$ \;
Pull $k \in \argmax_{k \in [K]} \theta_k$ \;
Update $p_k$ as the posterior distribution of $\mu_k$
}
\caption{\label{algo:thompson}Thompson sampling algorithm}
		\end{algorithm}

The Thompson sampling algorithm described in Algorithm~\ref{algo:thompson} adopts a Bayesian point of view. From some posterior distribution $\pmb{p}$ on the arm means $\pmb{\mu}$, it samples the vector $\btheta \sim \pmb{p}$ and pulls an arm maximising $\theta_k$. It then updates its posterior distribution using the observed reward, according to the Bayes rule.
\begin{theorem}[\citealt{kaufmann2012thompson}]\label{thm:TSbound}
There exists a function $f$ depending only on the means vector $\pmb{\mu}$ such that for every problem instance with Bernoulli reward distributions and $\varepsilon>0$, the regret of Thompson sampling algorithm is bounded as
\begin{equation*}
\singleregret(T) \leq (1+\varepsilon) \sum_{k:\mu_k < \mu_{(1)}} \frac{\mu_{(1)} - \mu_k}{\kl{\mu_k, \mu_{(1)}}}\log(T) + \frac{f(\pmb{\mu})}{\varepsilon^2}.
\end{equation*}
\end{theorem}
Despite coming from a Bayesian point of view, it thus reaches optimal frequentist performances, when initialized with a uniform prior. 
Sampling from the posterior distribution $\pmb{p}$ might be computationally expensive at each time step. Yet in special cases, \eg binary or Gaussian rewards, the posterior update is very simple. In the general case, a \textit{proxy} of the exact posterior can be used, by deriving results from the two specific aforementioned cases. 

\subsubsection{Explore-then-Commit Algorithm}

\begin{algorithm}[ht]
   \DontPrintSemicolon
\SetKwInOut{Input}{input}
$\mathcal{A} \gets [K]$ \tcp*{active arms}
\While{$\#\mathcal{A}> 1$}{
pull all arms in $\mathcal{A}$ once \;
\lFor{all $k\in \mathcal{A}$ such that $\hat{\mu}_k + B_k(T)\leq \max_{i \in \mathcal{A}} \hat{\mu}_i - B_i(T)$}{$\mathcal{A}\gets \mathcal{A}\setminus\{k\}$}
}
\lRepeat{$t=T$}{pull only arm in $\mathcal{A}$}
\caption{\label{algo:etc}Successive Eliminations algorithm}
		\end{algorithm}

While the above algorithms combine exploration and exploitation at each round, the ETC algorithm instead clearly separates both in two distinct phases. 
It first explores all the arms. Only once the best arm is detected (with high probability), it enters the exploitation phase and pulls this arm until the final horizon $T$.

Distinctly separating the exploration and the exploitation phase leads to a larger regret bound. Especially, if all the arms are explored the same amount of time (uniform exploration), the instance dependent bound scales with $\frac{1}{\Delta^2}$. 
Instead, the exploration can be adapted to each arm as described in Algorithm~\ref{algo:etc}. This finer version of ETC is referred to as Successive Eliminations \citep{perchet2013}.
An arm $k$ is \textit{eliminated} when it is detected as suboptimal, \ie when there is some arm $i$ such that $\hat{\mu}_k + B_k(T)\leq \hat{\mu}_i - B_i(T)$, for confidence bounds $B_i(T)$. When this condition holds, the arm~$k$ is worse than the arm $i$ with high probability; it is thus not pulled anymore.
With this adaptive exploration, the regret bound is optimal up to some constant factor as given by \cref{thm:etcbound}. 
\begin{theorem}[\citealt{perchet2013}]\label{thm:etcbound}
Algorithm~\ref{algo:etc} with $B_i(t) = \sqrt{\frac{2\log(T)}{N_i(t)}}$ has a regret bounded as
\vspace{-0.5em}
\begin{gather*}
\singleregret(T) =\bigO{\sum_{k: \mu_k<\mu_{(1)}} \frac{\log(T)}{\mu_{(1)}-\mu_k}},\\
\singleregret(T) = \bigO{\sqrt{KT\log(T)}}.
\end{gather*}
\end{theorem}
Besides yielding a larger regret than UCB and Thompson sampling of constant order, Successive Eliminations requires a prior knowledge of the horizon $T$. Knowing the horizon~$T$ is not too restrictive in bandits problem, since we can use the doubling trick \citep{degenne2016anytime}.
On the other hand, Successivation Eliminations has the advantage of being simpler to analyse (especially for more complex problems) as it clearly separates both exploration and exploitation.

\setcitestyle{numbers}
\section{Summary Tables}\label{app:table}

\cref{table1,table2} below summarize the theoretical guarantees of the algorithms presented in this survey. Unfortunately, some significant algorithms such as GoT \citep{bistritz2020b} are omitted, as the explicit dependencies of their upper bounds with other problem parameters than $T$ are unknown and not provided in the original papers. 

Algorithms using baselines different from the optimal matching in the regret definition are also omitted, as they can not be easily compared with other algorithms. This includes algorithms taking only a stable matching as baseline in the heterogeneous case, or algorithm which are robust to jammers for instance.

The upper bounds given in \cref{table1,table2} hide universal constant factors. Theoretical analyses of bandits algorithms often yield very large constant factors. These constants are often much larger than the ones observed in practice, because of looseness in the theoretical analysis. Because of these constant factors, some algorithms can outperform others in practice, despite having worse theoretical bounds.

\medskip

Here is a list of the different notations used in \cref{table1,table2}.
\begin{longtable}{p{.18\textwidth} p{.75\textwidth}}
$\Delta$ & $= \min\{U^*-U(\pi) > 0 \mid \pi \in \matchset\}$ \\
$\Delta_{(k,m)}$ & $= \min\{U^*-U(\pi) > 0 \mid \pi \in \matchset \text{ and } \pi(m)=k\}$ \\
$\bar{\Delta}_{(M)}$ & $= \min_{k\leq M} \mu_{(k)} - \mu_{(k+1)}$ \\
$\delta$ & arbitrarily small positive constant \\
$\mu_{(k)}$ & $k$-th largest mean (homogeneous case) \\
$M$ & number of players simultaneously in the game\\
$\matchset$ & set of matchings between arms and players \\
$N$ & total number of players entering/leaving the game (dynamic) \\
attackability & length of longest time sequence with successive $X_k(t)=0$ \\
rank & different ranks are attributed beforehand to players \\
$T$ & horizon \\
$U(\pi)$ & $= \sum_{m=1}^M \mu_{\pi(m)}^m$ \\
$U^*$ & $= \max_{\pi \in \matchset} U(\pi)$
\end{longtable}
\begin{table}[ht]
\begin{adjustwidth}{-4in}{-4in}
\centering
\scriptsize{{\setlength{\extrarowheight}{5pt}
\begin{tabular}{|c|c|c|c|c|}
\hline
\textbf{Model} & \textbf{Reference} & \textbf{Prior knowledge} 
& \textbf{Extra consideration}
& \textbf{Upper bound} \\[5pt]
\hline
Centralized & \mbox{\cucb \citep{chen2013}} & $M$
& -
&${\mathlarger \sum_{\substack{\pi \in \matchset \\ U(\pi) < U^*}} \frac{\log(T)}{U^* - U(\pi)}}$\\\hline
Centralized & \mbox{\textsc{CTS} \citep{wang2018}} & $M$
& Independent arms
&${\mathlarger \sum_{m=1}^M \mathlarger\sum_{k=1}^K \frac{\log^2(M)\log(T)}{\Delta_{(k,m)}}}$\\[10pt]\hline\hline
Coll. sensing & \mbox{$\text{dE}^3$ \citep{nayyar2016}} & $T, \Delta, M$ & communicating players &  $M^3 K^2\frac{\log(T)}{\Delta^2}$ \\[5pt] \hline
Coll. sensing & \mbox{\textsc{D-MUMAB}\citep{magesh2019a}} & $T, \Delta$ & unique optimal matching & $\frac{M\log(T)}{\Delta^2} + \frac{KM^3 \log(\frac{1}{\Delta})\log(T)}{\log(M)}$ \\[5pt] \hline
Coll. sensing & \mbox{ELIM-ETC \citep{boursier2019}} & $T$ & \makecell{$\delta=0$ if unique optimal matching}& $\mathlarger\sum_{k=1}^K \mathlarger\sum_{m=1}^M \left(\frac{M^2 \log(T)}{\Delta_{(k,m)}}\right)^{1+\delta}$ \\ \hline
Coll. sensing & \mbox{BEACON \citep{shi2021heterogeneous}} & - & - & $\mathlarger\sum_{k=1}^K \mathlarger\sum_{m=1}^M \frac{M \log(T)}{\Delta_{(k,m)}} + M^2K\log(K)\log(T)$ \\ \hline
\end{tabular}}}
\end{adjustwidth}
\caption{\label{table2} Summary of presented algorithms in the heterogeneous setting. The last column provides the asymptotic upper bound, up to universal multiplicative constant.}
\end{table}
\begin{table}[ht]
\begin{adjustwidth}{-4in}{-4in}
\centering
\scriptsize{{\setlength{\extrarowheight}{5pt}
\begin{tabular}{|c|c|c|c|c|}
\hline
\textbf{Model} & \textbf{Reference} & \textbf{Prior knowledge} 
& \textbf{Extra consideration}
& \textbf{Upper bound} \\[5pt]
\hline
Centralized & \mbox{\textsc{MP-TS} \citep{komiyama2015}}
& $M$ 
& -
&  ${\mathlarger \sum_{k > M}} \frac{\log(T)}{\mu_{(M)} - \mu_{(k)}}$ \\[8pt]\hline
\hline
Full sensing & \mbox{\textsc{SIC-GT} \citep{boursier2020}} & $T$ & robust to selfish players & ${\mathlarger\sum_{k > M}} \frac{\log(T)}{\mu_{(M)} - \mu_{(k)}} + MK^2 \log(T)$  \\\hline
\hline
 Stat. sensing & \mbox{\textsc{MCTopM} \citep{besson2018}} &$M$ & - & $M^3 \!\!\!\! {\mathlarger\sum_{1 \leq i < k \leq K}} \frac{ \log(T)}{\left(\mu_{(i)}- \mu_{(k)}\right)^2}$  \\
\hline 
 Stat. sensing & \mbox{\textsc{RR-SW-UCB\#} \citep{wei2018}} & $T, M$, rank & $\bigO{T^\nu}$ changes of $\pmb{\mu}$ & $\frac{K^2 M}{\Delta^2} T^{\frac{1+\nu}{2}} \log(T)$  \\[5pt]\hline 
 Stat. sensing & \makecell{\textsc{Selfish-Robust}\\ \textsc{MMAB} \citep{boursier2020}} & $T$ & robust to selfish players & $M{\mathlarger\sum_{k > M}} \frac{\log(T)}{\mu_{(M)} - \mu_{(k)}} + \frac{MK^3}{\mu_{(K)}} \log(T)$\\\hline
\hline
Coll. sensing & \mbox{\textsc{MEGA} \citep{avner2014}} & - & - & $M^2KT^{\frac{2}{3}}$ \\[5 pt] \hline
Coll. sensing & \mbox{\textsc{MC} \citep{rosenski2016}}
  & $T, \mu_{(M)} \! - \! \mu_{(M+1)}$
& -
& $\frac{MK \log(T)}{\left(\mu_{(M)} - \mu_{(M+1)}\right)^2}$ \\
\hline
Coll. sensing & \mbox{\sicmmab \citep{boursier2018}} & $T$
& -
& ${\mathlarger\sum_{k > M}} \frac{\log(T)}{\mu_{(M)} - \mu_{(k)}} + MK \log(T)$ \\
\hline
Coll. sensing & \mbox{\wangalgo \citep{wang2020}} & $T$ & -
& ${\mathlarger\sum_{k > M}} \frac{\log(T)}{\mu_{(M)} - \mu_{(k)}} $
\\ \hline 
Coll. sensing & \mbox{C\&P \citep{alatur2020}} & $T$ & Adversarial rewards & $K^{\frac{4}{3}}M^{\frac{2}{3}}\log(M)^{\frac{1}{3}}T^{\frac{2}{3}}$ \\ \hline
Coll. sensing & \mbox{\citep{bubeck2020a}} & $T$, rank, two players & Adversarial rewards & $K^2 \sqrt{T\log(K)\log(T)}$\\\hline\hline
No-sensing & \mbox{\citep{lugosi2018}} 
& $T, M$ 
& -
& $\frac{MK \log(T)}{\left(\mu_{(M)} - \mu_{(M+1)}\right)^2}$ \\
\hline
No-sensing & \mbox{\citep{lugosi2018}}
& $T, M, \mu_{(M)}$ 
& -
&$ \frac{MK^2}{\mu_{(M)}} \log^2(T) + MK \frac{\log(T)}{\Delta}$\\ \hline
No-sensing & \mbox{\citep{shi2020a}} & $T, \mu_{(K)}, \Delta$ & -
& ${\mathlarger\sum_{k > M}} \frac{\log(T)}{\mu_{(M)} - \mu_{(k)}} + M^2 K \frac{\log(\frac{1}{\Delta})\log(T)}{\mu_{(K)}}$\\ \hline
No-sensing & \mbox{\citep{huang2021}} & $T$ & - 
& ${\mathlarger\sum_{k > M}} \frac{\log(T)}{\mu_{(M)} - \mu_{(k)}} + MK^2  \log(\frac{1}{\Delta})^2\log(T)$\\ \hline
No-sensing & \mbox{A2C2 \citep{shi2020b}} & $T, M$, $\alpha$ & \makecell{Adversarial rewards \\ attackability $\bigO{T^\alpha}$} & $M^{\frac{4}{3}}K^{\frac{1}{3}}\log(K)^{\frac{2}{3}}T^{\frac{2+\alpha+\delta}{3}}$ \\\hline
No-sensing & \mbox{\citep{bubeck2020c}} & \makecell{$M$, rank\\shared randomness} & \makecell{No collision \\with high proba}
& $MK^{\frac{11}{2}}\sqrt{T\log(T)}$\\\hline
No-sensing & \mbox{\citep{bubeck2020a}} & $M$, rank & Adversarial rewards & $MK^{\frac{3}{2}}T^{1- \frac{1}{2M}}\sqrt{\log(K)}$
\\
\hline\hline
\makecell{No-sensing \\ Non zero collision $(M\geq K)$} & \mbox{\citep{bande2019a}} & $T, M, \Delta$ & Small variance of noise & $\frac{KM}{\Delta}e^{\frac{M-1}{K-1}}\log(T)$
\\\hline
\makecell{No-sensing \\ Non zero collision $(M\geq K)$} & \mbox{\citep{bande2021}} & $M,$ rank & - & $\frac{M^3K}{\Delta^2}\log(T)$
\\\hline
\hline
Dynamic, coll. sensing & \mbox{\textsc{DMC} \citep{rosenski2016}} & $T, \bar{\Delta}_{(M)}$ 
& -
&  $\frac{M \sqrt{K \log(T) T}}{\bar{\Delta}_{(M)}^2}$\\
\hline
Dynamic, coll. sensing & \mbox{\citep{bande2019a}} & $T$
& \makecell{Adversarial rewards \\ $N \leq \bigO{\sqrt{T}}$}
& $\frac{K^{K+2}}{\sqrt{K\log(K)}}T^{\frac{3}{4}}+NK\sqrt{T}$  \\\hline
Dynamic, no-sensing & \mbox{\textsc{DYN-MMAB} \citep{boursier2018}} & $T$
& All players end at $T$
&  $  \frac{MK \log(T)}{\bar{\Delta}_{(M)}^2}+ \frac{M^2K \log(T)}{\mu_{(M)}} $ \\\hline
\end{tabular}}}
\end{adjustwidth}
\caption{\label{table1} Summary of presented algorithms in the homogeneous setting. The last column provides the asymptotic upper bound, up to universal multiplicative constant.}
\end{table}

\FloatBarrier

\newpage
\bibliography{multiplayer}
\end{document}